\documentclass{article}

    \PassOptionsToPackage{numbers, compress}{natbib}
\usepackage[preprint]{neurips_2026}

\usepackage[utf8]{inputenc} 
\usepackage[T1]{fontenc}    
\usepackage{hyperref}       
\usepackage{url}            
\usepackage{booktabs}       
\usepackage{amsfonts}       
\usepackage{nicefrac}       
\usepackage{microtype}      
\usepackage{xcolor}         
\usepackage{nicematrix}
\usepackage{marvosym}
\newcommand{\name}{{\sc SPA }}
\newcommand{\mame}{{\sc SPA}}

\usepackage{subcaption}
\usepackage{multirow}
\usepackage{amsmath,amssymb}
\usepackage{mathtools}
\usepackage{xfrac}
\usepackage[english]{babel}
\usepackage{hyphenat}
\usepackage{nicematrix}
\usepackage{tabularx} 
\usepackage{xcolor} 
\usepackage{colortbl}
\usepackage{soul}
\usepackage{csquotes}
\usepackage{listings}

\usepackage{seqsplit}
\usepackage{arydshln}
\usepackage{algorithm}
\usepackage{algorithmic}

\newcommand{\R}{\mathbb{R}}

\usepackage{xspace}

\usepackage{wrapfig}
\definecolor{Gray}{gray}{0.85}
\definecolor{Redo}{rgb}{0.95,0.69,0.51}
\definecolor{LightCyan}{rgb}{0.88,1,1}
\usepackage{afterpage}
\usepackage{pifont}
\usepackage{csquotes}
\usepackage{lipsum}

\title{Unlocking Patch-Level Features \\ for CLIP-Based Class-Incremental Learning}

\author{Hao Sun $^{1,2}$  \ \ \ Zi-Jun Ding $^{1,2}$ \ \ \
Da-Wei Zhou$^{1,2}\textsuperscript{(\Letter)}$ \\
$^{1} $ School of Artificial Intelligence, Nanjing University \\
$^{2} $ State Key Laboratory for Novel Software Technology, Nanjing University \\
\texttt{\small sunhao@lamda.nju.edu.cn, dingzijun@njust.edu.cn, zhoudw@lamda.nju.edu.cn}\\
}
\begin{document}
\maketitle

\begin{abstract}
 Class-Incremental Learning (CIL) enables models to continuously integrate new knowledge while mitigating catastrophic forgetting. Driven by the remarkable generalization of CLIP, leveraging pre-trained vision-language models has become a dominant paradigm in CIL. However, current work primarily focuses on aligning global image embeddings (\textit{i.e.}, [CLS] token) with their corresponding text prompts (\textit{i.e.}, [EOS] token). Despite their good performance, we find that they discard the rich patch-level semantic information inherent in CLIP's encoders. For instance, when recognizing a \textit{rabbit}, local patches may encode its distinctive cues, such as long ears and a fluffy tail, which can provide complementary evidence for recognition. Based on the above observation, we propose \name (\underline{\textbf{S}}emantic-guided \underline{\textbf{P}}atch-level \underline{\textbf{A}}lignment) for CLIP-based CIL, which aims to awaken long-neglected local representations within CLIP. Specifically, for each class, we first construct representative and diverse visual samples and feed them to GPT-5 as visual guidance to generate class-wise semantic descriptions. These descriptions are used to guide the selection of discriminative patch-level visual features. Building upon these selected patches, we further employ optimal transport to align selected patch tokens with semantic tokens from class-wise descriptions, yielding a structured cross-modal alignment that improves recognition. Furthermore, we introduce task-specific projectors for effective adaptation to downstream incremental tasks, and sample pseudo-features from stored class-wise Gaussian statistics to calibrate old-class representations, thereby mitigating catastrophic forgetting. Extensive experiments demonstrate that \name achieves state-of-the-art performance.
\end{abstract}    
\section{Introduction}\label{sec1}
In recent years, the rapid development of deep learning~\citep{he2016deep, hendrycks2021many,liu2015deep} has profoundly impacted numerous fields. However, real-world scenarios are inherently dynamic, with data often appearing as continuous and non-stationary streams~\citep{rebuffi2017icarl, zhao2020maintaining}. In such contexts, traditional deep learning models can suffer from catastrophic forgetting~\citep{french1999catastrophic,serra2018overcoming,shi2021overcoming}, as they tend to overwrite previously learned knowledge when trained on shifting data distributions. To address this issue, Class-Incremental Learning (CIL)~\citep{de2021continual,gao2022r,zhou2024class,zhou2025revisiting} has emerged as an essential paradigm that enables models to incrementally absorb new concepts while retaining previously acquired knowledge. Recently, the rapid advancement of pre-trained models~\citep{dosovitskiy2020image, li2023blip}, particularly Vision-Language Models (VLMs)~\citep{yao2022pevl}, like CLIP~\citep{radford2021learning}, has significantly accelerated a paradigm shift in the CIL~\citep{wang2022learning,zhou2024continual}.
By harnessing their powerful generalization capabilities, current research~\citep{huang2025mind,zhou2025learning} has shifted from training networks from scratch to applying Parameter-Efficient Fine-Tuning (PEFT) strategies~\citep{hu2025hierarchical,wang2022learning} on frozen PTM backbone networks, requiring the adjustment of only a small number of additional  parameters~\citep{jia2021scaling}.

CLIP~\citep{radford2021learning} leverages a contrastive learning paradigm to project images and texts into a shared embedding space. Its powerful zero-shot generalization and representation learning capabilities allow it to effectively handle various downstream tasks. Specifically, its image encoder prepends a learnable [CLS] token to the input sequence of image patches to aggregate global visual features, while the text encoder relies on the [EOS] token to capture global textual semantics. Although both encoders naturally generate rich patch-token and word-token sequences during the encoding process~\citep{wu2024llm2clip}, the CLIP model only computes the cosine similarity between the global [CLS] and [EOS] embeddings for classification. 

\begin{figure*}[t]
	\vspace{-7mm}
	\begin{center}
		{\includegraphics[width=0.95\textwidth]{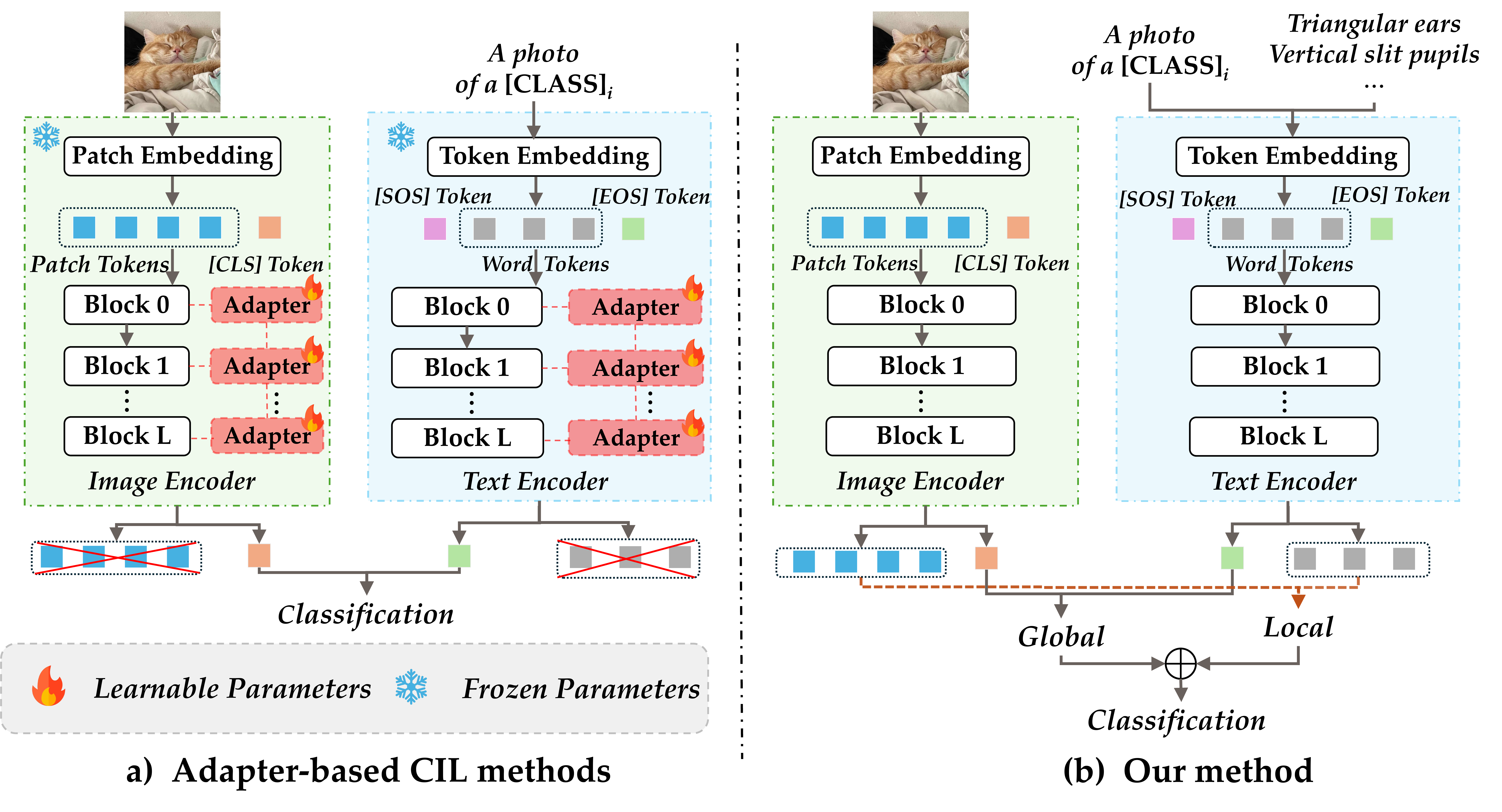}}
	\end{center}
	\vspace{-3mm}
	\caption{  Comparison of CLIP-based CIL paradigms.  (a) Existing adapter-based CIL methods mainly rely on the global [CLS]
 and [EOS] tokens for cross-modal alignment, leaving the rich semantic information in patch-level tokens insufficiently explored. (b) Our method explicitly aligns patch-level tokens, activating the previously overlooked local features and improving CIL performance.
	}
	\vspace{-5mm}
	\label{figure:pre}
\end{figure*}

Current CLIP-based CIL research~\citep{huang2025mind} primarily focuses on PEFT strategies (\textit{e.g.}, via adapters or prompt tuning)~\citep{gao2024clip,wu2024controlmllm,yu2024boosting,zhou2022conditional}, as illustrated in Figure~\ref{figure:pre}(a). While these methods have improved CIL performance, they also inherently inherit the global alignment paradigm of the original CLIP model, resulting in the underutilization of rich local visual features and semantic contexts. For instance, when recognizing a \textit{rabbit}, different local patches may encode distinctive visual cues such as long ears, a short fluffy tail, or soft fur texture. Explicitly leveraging these patch-tokens can provide additional evidence for recognition.

Awakening and utilizing the local information carried by token-level image-text offers a promising approach to enhance the discriminative capabilities in CIL, but it also presents two significant challenges: \textbf{1)} The original image patch-tokens are often filled with a considerable amount of background noise~\citep{chen2022plot}, and simply extracting all the patch information without proper filtering can interfere with classification tasks by introducing irrelevant features. This makes it difficult for the model to effectively leverage the rich spatial details embedded within the patches, leading to suboptimal performance. \textbf{2)} Directly aligning local patches with global text prompts (\textit{e.g.}, ``a photo of a [CLASS]") suffers from a semantic level mismatch. The class prompt encodes global semantics, whereas patch features correspond to local regions. Forcing these local visual representations to match the same global  semantics ignores their semantic diversity, making different patches less distinguishable and weakening the effectiveness of patch-level features in classification.

To address the above challenges, we propose \name (\underline{\textbf{S}}emantic-guided \underline{\textbf{P}}atch-level \underline{\textbf{A}}lignment) for CLIP-based CIL. Different from existing methods, \name explicitly activates the long-neglected patch-level representations within CLIP and leverages them to improve performance in CIL. Specifically, we first construct representative and diverse visual samples for each class, and then leverage GPT-5 to generate class-wise attribute semantics from these samples. Based on these semantics, \name evaluates the semantic relevance of image patches and selects the top-$K$ most discriminative patch-level visual features. We then align these selected patch-level features with multiple class-wise semantic embeddings via optimal transport, thereby achieving structured cross-modal alignment. 
To mitigate catastrophic forgetting, we further introduce expandable task-specific projectors and Gaussian pseudo-feature sampling. The projectors support task-wise adaptation by updating only newly added modules, while the calibration module samples pseudo-features from stored class-wise statistics to preserve old-class distributions. 
Experimental results show that \name significantly improves the discriminative power of CLIP in CIL, surpassing existing methods across nine benchmark datasets.

\section{Related Work}
\noindent\textbf{Class-Incremental Learning.}
CIL aims to enable models to continuously learn new classes from a data stream while mitigating catastrophic forgetting~\citep{de2021continual,masana2022class}. Traditional CIL methods can be broadly categorized into several groups. Regularization-based methods~\citep{aljundi2018memory,kirkpatrick2017overcoming,zenke2017continual} impose constraints on key updated parameters, penalizing changes in the critical parameter space.
Replay-based methods mitigate forgetting by preserving a memory buffer~\citep{chaudhry2018efficient,luo2023class} of exemplar samples from previous tasks or by employing generative models~\citep{ostapenko2019learning,xiang2019incremental} to reconstruct and replay prior data distributions. Dynamic network-based methods expand the network structure to accommodate new tasks, including neuron expansion~\citep{xu2018reinforced,yoon2017lifelong}, backbone expansion~\citep{wang2022beef, zheng2025task}, and prompt expansion~\citep{liu2021adaptive}. Knowledge distillation-based methods~\citep{hinton2015distilling,li2017learning} transfer knowledge from previously trained models to the current one, ensuring that new tasks are learned while retaining important information from earlier tasks.\\
\noindent\textbf{Pre-Trained Model-Based CIL.} With the rapid development of PTMs, such as ViT~\citep{dosovitskiy2020image} and CLIP~\citep{radford2021learning}, the research focus of CIL has gradually shifted from conventional training with random initialization to an efficient adaptation paradigm based on PTMs~\citep{li2025addressing,qi2025adaptive}. To adapt to downstream tasks while preserving the generalization capability, existing ViT-based methods typically freeze the pre-trained backbone and introduce lightweight learnable modules~\citep{smith2023coda,wang2022s}, such as learnable prompts prepended to the input token sequence~\citep{wang2022learning} or adapters inserted into selected layers of the Transformer backbone~\citep{fukuda2025adapter,yu2024boosting}. CLIP-based methods~\citep{zhou2025learning} further exploit the cross-modal alignment capability of vision-language models to improve CIL performance. RAPF~\citep{huang2024class} adaptively calibrates old-class representations and alleviates catastrophic forgetting through decomposed parameter fusion after adapter tuning. CLG-CBM~\citep{yu2025language} builds a language-guided concept bottleneck model, leveraging interpretable concept representations to improve CIL performance while enhancing model interpretability.

\section{Preliminaries}

\textbf{Class-Incremental Learning.}
Formally, CIL aims to endow models with the ability to continuously acquire new knowledge from $B$ distinct incremental tasks, denoted as $\left\{\mathcal{D}^{1}, \mathcal{D}^{2}, \cdots, \mathcal{D}^{B}\right\}$, while retaining previously learned information~\citep{rebuffi2017icarl}. At the  $b$-th task, the model is trained on a dataset $\mathcal{D}^{b}=\left\{\left(\mathbf{x}_{i}, y_{i}\right)\right\}_{i=1}^{n_b}$ consisting of $n_b$ examples, where $\mathbf{x}_{i}\in  \R^D$  represents an instance of class $y_i\in Y_b$, with $Y_b$ denoting the label space corresponding to task $b$. The label spaces of different tasks are disjoint (\textit{i.e.}, $Y_b  \cap Y_{b^\prime} = \varnothing$ for $b\neq b^\prime$). In this paper, we adopt the exemplar-free protocol~\citep{wang2022dualprompt,zhou2022learning}, where at each task $b$, the model has access only to the training data of the current task $\mathcal{D}^{b}$, and cannot access any data from previous tasks  $\mathcal{D}^{1:b-1}$. The goal of CIL is to build a unified classifier for all seen classes $\mathcal{Y}_b=Y_1 \cup \cdots \cup Y_b$. Formally, we aim to learn a model $f(\mathbf{x}): X\rightarrow\mathcal{Y}_b$ that minimizes the expected risk:
\begin{equation} \label{eq:totalrisk} 
 	f^*=\underset{f \in \mathcal{H}}{\operatorname{argmin}} \; \mathbb{E}_{(\mathbf{x}, y) \sim \mathcal{D}_{t}^1\cup\cdots\mathcal{D}_{t}^b} \mathbb{I}\left(y \neq f(\mathbf{x})\right) \, , 
 \end{equation}
where $\mathcal{H}$ represents the hypothesis space, $\mathbb{I}(\cdot)$  denotes the indicator function, and $\mathcal{D}_{t}^b$ refers to the data distribution of the $b$-th task.\\
\textbf{Vision-Language Model.} Following~\citep{huang2024class,zhou2025learning}, we use the pre-trained model CLIP ~\cite{radford2021learning} as the initialization for $f(\mathbf{x})$. CLIP consists of an image encoder $g_i(\cdot)$ and a text encoder $g_t(\cdot)$, which together project images and texts into a shared $d$-dimensional embedding space. Given an input image $\mathbf{x} \in \R^{D}$, the image encoder $g_i(\cdot)$ (\textit{e.g.}, Vision Transformer~\citep{dosovitskiy2020image}) first divides it into $M$ non-overlapping image patches. These image patches are linearly projected, and positional encodings are added to form the image embeddings. Subsequently, a learnable class token (\textit{i.e.}, [CLS] token) is prepended to the image embeddings, forming the input sequence for the Transformer layers. Then, the [CLS] token aggregates global information from the patch embeddings through interactions and the multi-head self-attention mechanism, ultimately serving as the global visual representation of the image. Meanwhile, each patch embedding interacts with other patch-tokens to enrich its own representation, ultimately yielding the feature embedding sequence $\mathbf{V} = [\mathbf{v}_\text{cls},\mathbf{v}_1,\cdots,\mathbf{v}_M],$
where $ \mathbf{v}_\text{cls} = g_i(\mathbf{x})\in \R^d$ is the global visual representation, and $\{\mathbf{v}_i\}_{i=1}^M$ corresponds to the patch-tokens containing rich local information that remains underutilized. Similarly, for a given class $y_i\in \mathcal{Y}_b$, we construct a templated prompt $\mathbf{c}_{i}~$(\textit{e.g.}, ``a photo of a [CLASS]''). The text encoder $g_t(\cdot)$ tokenizes the prompt, prepends a start token (\textit{i.e.}, [SOS] token) and appends an end token (\textit{i.e.}, [EOS] token), encoding it into the sequence of embeddings $\mathbf{T}^i =  [\mathbf{t}_{\text{sos}}^i,\mathbf{t}_1^i,\cdots,\mathbf{t}_{N}^i,\mathbf{t}_{\text{eos}}^i],$ where $ \mathbf{t}_\text{eos}^i =g_t(\mathbf{c}_{i})\in \R^d$ serves as the global semantic representation.
During the inference phase, CLIP relies on computing the similarity between the global visual feature $\mathbf{v}_\text{cls}$  and the global text embeddings $\mathbf{t}_\text{eos}$ of each class in $\mathcal{Y}_b$ for category prediction. The probability that the image 
$\mathbf{x}$ belongs to class $y_i$ is calculated as:
 \begin{equation} \label{eq:clip_pred}
	f_{y_i}(\mathbf{x}, \mathbf{c}_{i})  =\frac{\exp \left(\cos \left(\mathbf{v}_\text{cls}, \mathbf{t}_\text{eos}^{i}  \right) / \tau\right)}{\sum_{j=1}^{|\mathcal{Y}_b|} \exp \left(\cos \left(\mathbf{v}_\text{cls}, \mathbf{t}_\text{eos}^{j} \right) / \tau\right)} =\frac{\exp \left(\cos \left(g_i(\mathbf{x}), g_t(\mathbf{c}_i)  \right) / \tau\right)}{\sum_{j=1}^{|\mathcal{Y}_b|} \exp \left(\cos \left(g_i(\mathbf{x}), g_t(\mathbf{c}_j) \right) / \tau\right)} ,
\end{equation}
where $\cos(\cdot,\cdot)$ represents the cosine similarity and $\tau$ denotes the temperature parameter. In Eq.~\ref{eq:clip_pred}, cross-modal alignment relies solely on [CLS] token $\mathbf{v}_\text{cls}$ and [EOS] token $\mathbf{t}_\text{eos}$, while the rich local visual patch-token features $\{\mathbf{v}_i\}_{i=1}^M$  and textual token-level semantics $\{\mathbf{t}_j^i\}_{j=1}^N$  are ignored.\\
\textbf{Baselines in Class-Incremental Learning.} Existing CIL methods using pre-trained models primarily rely on PEFT strategies to adapt to new tasks while mitigating catastrophic forgetting. Representative approaches include prompt-based methods~\citep{wang2022s,wang2022dualprompt,wang2022learning}, which introduce a learnable prompt pool $\mathcal{P}$, and adapter-based methods~\citep{fukuda2025adapter,gao2024clip,tan2024semantically}, which insert lightweight trainable adapters $\mathbf{M}_v$ into the frozen Transformer backbone. Both paradigms refine the global representations (\textit{e.g.}, $\tilde{\mathbf{v}}_\text{cls}$) using only a small number of learnable parameters, enabling efficient adaptation without updating the backbone.\\
\textbf{Discussions.}  
Although these paradigms achieve efficient adaptation to downstream tasks and mitigate catastrophic forgetting, they primarily inherit CLIP's global alignment mechanism in Eq.~\ref{eq:clip_pred}, leaving the richly informative patch-level features and token-level textual semantics underexplored. Therefore, explicitly exploiting these local features is crucial for enhancing recognition.

\section{\mame: Semantic-guided Patch-level Alignment} 
\begin{figure*}[t]
	\vspace{-7mm}
	\begin{center}       	{\includegraphics[width=0.95\columnwidth]{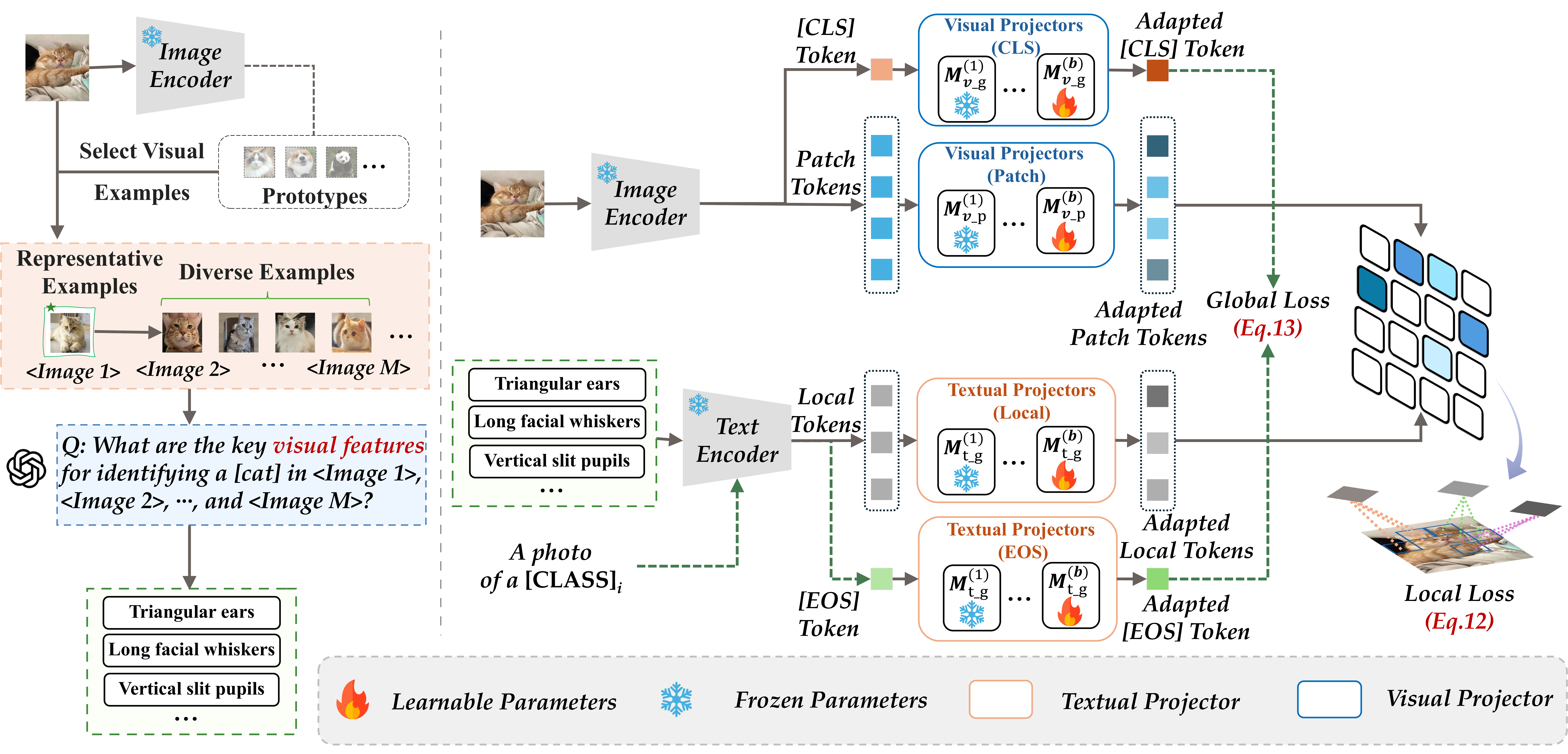}}
	\end{center}
	\vspace{-2mm}
	\caption{  Illustration of \mame. \textbf{Left:} For each class, \name first constructs representative and diverse visual samples and uses GPT-5 to generate class-wise semantics. \textbf{Right:} These semantics are then used to select discriminative image patches, and \name performs structured local alignment via optimal transport with task-specific projectors to mitigate catastrophic forgetting.
	}
	\vspace{-5mm}
	\label{figure:teaser}
\end{figure*}
Motivated by the underexplored local visual and textual semantics in CLIP-based CIL, we propose \mame, a semantic-guided patch-level alignment framework. As shown in Fig.~\ref{figure:teaser}, \name first selects representative and diverse samples as visual references, and leverages GPT-5 to generate class-wise attribute semantics. Guided by these semantics, \name performs
discriminative patch selection and employs optimal transport to achieve structured alignment between selected visual tokens and textual attribute tokens. To mitigate forgetting, we further introduce task-specific visual and textual projectors, together with Gaussian pseudo-feature sampling for preserving old-class distributions.
\subsection{ Class-wise Attribute Semantic Generation}\label{sec:4.1}
Most prior CLIP-based CIL methods construct text prompts solely from class names (\textit{e.g.}, ``A photo of a $\text{[CLASS]}_i$''). These prompts are encoded into global text representations, and classification is then performed following Eq.~\ref{eq:clip_pred}. However, such prompts cannot fully capture visual attributes such as color and local structure, which often play a crucial role in distinguishing categories. To provide richer textual priors, we construct class-wise attribute semantics for subsequent discriminative patch selection and local cross-modal alignment.\\
\textbf{Visual Sample Construction.} To generate more discriminative class-wise attribute semantics, we first construct a visual sample set for each class that is both representative and diverse. Specifically, for each class $c \in Y_b$ at task $b$, we first compute its visual prototype $\mathbf{p}_c$ based on the global features extracted by the frozen visual encoder $\bar{g_i}(\cdot)$:
\begin{align}
\label{eq:prototype}
\mathbf{p}_c = \frac{1}{N} \sum_{j=1}^{|\mathcal{D}^{b}|} \mathbb{I}(y_j=c)\, \bar{g_i}(\mathbf{x}_j),
\end{align}
where $N =\sum_{j=1}^{|\mathcal{D}^{b}|} \mathbb{I}(y_j=c)\ $ denotes the number of samples belonging to class $c$, $\mathbb{I}(\cdot)$  denotes the indicator function. We define the cosine distance between two features $\mathbf{a}$ and $\mathbf{b}$ as  $d \left(\mathbf{a},\mathbf{b}\right) = 1 -\frac{\mathbf{a}^\top\mathbf{b}}{\|\mathbf{a}\|_2\|\mathbf{b}\|_2}$,
where $\|\cdot\|_2$ denotes the L2-norm. Based on this prototype, we select the sample whose global feature is closest to the prototype $\mathbf{p}_c$ as the representative sample $\mathbf{x}_{c}^r$:
\begin{align}
\label{eq:vis_pro}
\mathbf{x}_{c}^r = \arg \underset{\mathbf{x}_j \in \mathcal{D}^{b},\, y_j=c}{\min} d \left(\bar{g_i}\left(\mathbf{x}_j\right),\mathbf{p}_c\right),
\end{align}
 However, relying on a single representative sample is insufficient to fully capture intra-class visual variations, and may introduce instance-specific bias into subsequent semantic construction. To enrich the diversity of the class-level visual context, we further select $n$ samples that are most dissimilar to $\mathbf{x}_{c}^r$ from the remaining samples of the same class.
The final visual sample set is constructed as: 
 \begin{align}
\label{eq:vis_other}
\mathcal{S}_c = \{\mathbf{x}_{c}^r\}  \cup \mathrm{Topn}_{\mathbf{x}_j \in \mathcal{D}^{b},y_j=c,\,\mathbf{x}_j \neq \mathbf{x}_{c}^r}  d \left(\bar{g_i}\left(\mathbf{x}_j\right),\bar{g_i}\left(\mathbf{x}_{c}^r\right)\right),
\end{align}
where $\mathrm{Topn}(\cdot)$ selects the top-$n$ samples with the largest distances. In this way, $\mathbf{x}_{c}^r$ provides representative visual information, while the additional $n$ samples capture the intra-class diversity. They jointly constitute the class-level visual context used for subsequent semantic generation.\\
\textbf{Attribute Semantic Generation.} Using the constructed class-level visual samples, we utilize GPT-5.4~\citep{achiam2023gpt} to generate attribute descriptions for each class. Specifically, for class $c$, its sample images and class name are fed into GPT-5.4 with the following prompt template to generate attribute semantics:
\begin{displayquote}
\vspace{-3mm}
\itshape
{\bf Input:}
<Image 1>, <Image 2>, $\cdots$ <Image M>, and [CLASS]$_c$\\
	 {\bf Q}: What are the key visual features for identifying a [CLASS]$_c$ in these images? Focus on the most discriminative attributes.\\
	 {\bf A}: {\bf 1.} Pointed triangular ears. {\bf 2.} Vertical-slit pupils. {\bf 3.}$\cdots$
   \vspace{-3mm}  
\end{displayquote}
Based on the above prompt, we generate a set of discriminative visual attributes for class $c$, denoted as $\mathrm{A}_c =\{a_{j}^c\}_{j=1}^{N_a}$, where $a_{j}^c$ denotes the $j$-th attribute of class $c$.
\subsection{Semantic-guided Discriminative Patch Selection}\label{sec:4.2}
The constructed attribute semantics provide class-specific textual priors for local representation learning. However, directly aligning them with all patch tokens may introduce background regions and class-irrelevant local patterns~\citep{chen2022plot}. For example, when recognizing a \textit{rabbit} on grass, background grass patches may be incorrectly treated as rabbit-related regions, which introduce noise and interfere with local representation learning. To reduce such interference, we adopt semantic-guided discriminative patch selection to filter out class-irrelevant patches and preserve informative local patches.\\
\textbf{Discriminative Patch-level Set Selection.} Given an input image $\mathbf{x}$, we use the visual encoder $\bar{g_i}(\cdot)$ to extract the visual token sequence $[\mathbf{v}_\text{cls},\mathbf{v}_1,\cdots,\mathbf{v}_M]$. We then retain the patch tokens to form the patch-level feature set $\mathcal{V} =\{\mathbf{v}_i\}_{i=1}^M$. For each class $c$, we use the generated attribute descriptions $\mathrm{A}_c$ as class-wise semantic guidance. We randomly sample $N$ descriptions from $\mathrm{A}_c$, and encode them with the text encoder $\bar{g_t}(\cdot)$ to obtain the corresponding attribute embeddings $\mathcal{T}_c =\{\mathbf{t}_{n}^c\}_{n=1}^N$. We then compute the similarity between each patch-level feature and each attribute embedding of class $c$ as:
\begin{align}
\label{eq:similarity}
\text{sim}_{m,n}^c= \frac{\mathbf{v}_m^\top\mathbf{t}_{n}^c}{\|\mathbf{v}_m\|_2\|\mathbf{t}_{n}^c\|_2},
\end{align}
where $\text{sim}_{m,n}^c$ measures the semantic relevance between the $m$-th patch in image $\mathbf{x}$ and the $n$-th attribute of class $c$. Based on these similarities, we further aggregate the responses of each image patch to all attribute semantics of class $c$, and define the class-aware discriminative score as $\text{q}_{m}^c = \frac{1}{N}\sum_{n=1}^{N}\text{sim}_{m,n}^c.$ 
According to $\text{q}_{m}^c$, we select the top-$K$ patches from image $\mathbf{x}$ that are most relevant to the attribute semantics of class $c$, and denote the discriminative patch set from image $\mathbf{x}$ as:
\begin{align}
\label{eq:dis_set}
\mathcal{R}_c  =  \big\{ \mathbf{v}_m \,\big|\, \text{q}_{m}^c \in \mathrm{TopK}\big(\{ \text{q}_{m}^c \}_{m=1}^M \big) \big\},
\end{align}
where $\mathrm{TopK}(\cdot)$ denotes the operation of selecting the $K$ highest discriminative scores among all $M$ patches. Through the above selection process, the model is encouraged to preserve class-relevant local features and reduce interference from background and irrelevant regions.\\
\textbf{Task-specific Projector Construction.} Although CLIP exhibits strong zero-shot generalization, the domain gap between pre-training data and downstream incremental tasks makes adaptation necessary. Therefore, we introduce lightweight projectors after the frozen visual and textual encoders to adapt CLIP representations to downstream incremental tasks. However, using a single shared projector across all tasks may limit adaptation to new classes, and increase interference with previously learned knowledge. To address this issue, we introduce task-specific projectors to enhance task-wise adaptation while alleviating interference with previously acquired knowledge. Specifically, for the 
$b$-th task, we initialize a new task-specific visual projector $\mathbf{M}_v^b(\cdot):\R^d\rightarrow\R^d$  and a new task-specific textual projector 
$\mathbf{M}_t^b(\cdot):\R^d\rightarrow\R^d$ after the frozen visual and textual encoders, respectively, while keeping all projectors $\{\mathbf{M}_v^{1:b-1},\mathbf{M}_t^{1:b-1}\}$ learned from previous tasks frozen. Given a patch feature $\mathbf{v}_k$ and an attribute embedding $\mathbf{t}_n^c$, we obtain the adapted visual representation $\tilde{\mathbf{v}}_k$ and textual representation $ \tilde{\mathbf{t}}_n^c$ by cumulatively aggregating the outputs of the current and historical projectors:
\begin{align}
\label{eq:12}
\tilde{\mathbf{v}}_k = {\mathbf{M}}_v^b(\mathbf{v}_k) +\sum_{i=1}^{b-1}  \bar{\mathbf{M}}_v^i(\mathbf{v}_k), \quad
        \tilde{\mathbf{t}}_n^c = {\mathbf{M}}_t^b(\mathbf{t}_n^c) +\sum_{i=1}^{b-1} \bar{\mathbf{M}}_t^i(\mathbf{t}_n^c),
\end{align}
where $\bar{\mathbf{M}}_v^i(\cdot)$ and 
$\bar{\mathbf{M}}_t^i(\cdot)$ denote the frozen visual and textual projectors learned from task $i$, respectively. In this way, the current task can be adapted by newly introduced projectors, while the knowledge acquired from previous tasks is preserved through the frozen historical projectors. The aggregated features are then used to replace the raw visual and textual features in Eq.~\ref{eq:similarity}, and are further fed into the discriminative patch-level set selection and local alignment.
\subsection{Optimal Transport-based Patch-level Alignment}\label{sec:4.3}
After obtaining the discriminative patch set $\mathcal{R}_c$ in Sec.~\ref{sec:4.2}, directly matching the selected patches to textual attribute embeddings remains suboptimal. Naive matching strategies~\citep{lafon2024gallop} tend to force multiple distinct patches to concentrate on the same attribute semantics, leading to degenerate local alignment and feature redundancy. This weakens the stability of representation learning and aggravates forgetting of previously acquired knowledge. Therefore, it is essential to establish a balanced local alignment for robust representation learning in CIL.\\
\textbf{Alignment via Optimal Transport.} To address this issue, we formulate patch-level alignment as a joint assignment problem. By imposing marginal constraints, optimal transport~\cite{gangbo1996geometry} distributes the matching mass across patches and attributes, leading to more balanced local correspondences. Specifically, for class $c$, we rewrite the selected discriminative patch set from the input image as an ordered set $\mathcal{R}_c  =  \{\mathbf{r}_{k}^c\}_{k=1}^K$, where $\mathbf{r}_{k}^c\in \mathbb{R}^d$ denotes the feature of the $k$-th selected patch. The corresponding attribute embedding set is denoted as $\mathcal{T}_c =\{\mathbf{t}_{n}^c\}_{n=1}^N$. We first compute the pairwise semantic similarity $\text{sim}_{k,n} $ between each selected patch $\mathbf{r}_{k}^c$ and each attribute embedding  $\mathbf{t}_{n}^c$ using Eq.~\ref{eq:similarity}.
 Based on this similarity matrix, the transport cost 
matrix $\mathbf{C}_c \in \mathbb{R}^{K \times N}$ is defined as:
\begin{align}
\label{eq:transport cost matrix}
\mathbf{C}_c(k,n)  = 1-\text{sim}_{k,n},
\end{align}
To enforce balanced matching, we adopt uniform marginal distributions over the selected patches and attribute embeddings, \text{i.e.},  $\mathbf{a} = \frac{1}{K}\mathbf{1}_K,\mathbf{b} = \frac{1}{N}\mathbf{1}_N$. Therefore, the entropy-regularized balanced optimal transport problem is formulated and solved via the Sinkhorn iterations~\citep{cuturi2013sinkhorn} as follows:
\begin{align}
\mathbf{\Pi}_c^* = \arg\min_{\mathbf{\Pi} } \langle \mathbf{\Pi}, \mathbf{C}_c \rangle - \lambda \mathcal{H}(\mathbf{\Pi}) \quad \text{s.t.} \quad \mathbf{\Pi}\mathbf{1}_N = \mathbf{a}, \; \mathbf{\Pi}^\mathrm{T} \mathbf{1}_K = \mathbf{b},
\label{eq:transport}
\end{align}
where $\mathcal{H}(\cdot)$ denotes the entropy regularization term, and $\lambda > 0$ is a hyperparameter.
After obtaining the optimal transport matrix $\mathbf{\Pi}_c^*$, we define the local alignment score for class $c$ as the transport-weighted semantic similarity between the selected patches and the attribute embeddings:
\begin{align}
\label{eq:local alignment score}
\sigma^{\text{loc}}_{c} = \sum_{k=1}^{K} \sum_{n=1}^{N} \mathbf{\Pi}_c^{*}(k,n) \, \text{sim}_{k,n}.
\end{align}
We regard these scores of all seen classes as the local classification logits and apply softmax 
normalization to obtain $f_{\text{local},c}(\mathbf{x})$. The local loss is defined as:
\begin{align} 
	\mathcal{L}_{{l}} &= \ell (f_\text{local}(\mathbf{x}), y), \quad \text{where} \quad 
    f_{\text{local},c}(\mathbf{x})
=\frac{\exp{(\sigma^{\text{loc}}_{c}})}{\sum_{c'=1}^C \exp{(\sigma^{\text{loc}}_{c'}})}.\label{eq:local-loss}
\end{align}

\textbf{Global Loss.} To preserve stable global cross-modal alignment during incremental learning, we extract the global [CLS] token and the global [EOS] token from the frozen CLIP encoders, and transform them with a separate set of task-specific projectors for global semantic alignment, yielding the adapted global visual representation  $\tilde{\mathbf{v}}_\text{cls}$ and textual representation  $\tilde{\mathbf{t}}^c_\text{eos}$ using Eq.~\ref{eq:12}. Although task-specific projectors mitigate parameter interference across incremental tasks, the model may still be biased toward new classes. To mitigate this bias, we adopt Gaussian feature calibration~\cite{zhu2021prototype}. For class $c$, we store only its class-wise global statistics, \textit{i.e.}, the visual prototype $\mathbf{p}_c$ and covariance matrix $\mathbf{\Sigma}_c$. When learning task $b$, we sample old-class pseudo features from $\mathcal{N}(\mathbf{p}_c,\mathbf{\Sigma}_c)$, and combine them with current-task samples from $\mathcal{D}^b$ to optimize the global classification loss:
 \begin{align} \label{eq:unit-loss}
	\mathcal{L}_{g} = \ell (f_\text{global}(\mathbf{x}), y), \quad \text{where} \quad 
	f_{\text{global},{c}}(\mathbf{x})  =\frac{\exp(\cos(\tilde{\mathbf{v}}_\text{cls}, \tilde{\mathbf{t}}^c_\text{eos})/\tau)}{\sum_{j=1}^C \exp(\cos(\tilde{\mathbf{v}}_\text{cls}, \tilde{\mathbf{t}}^j_\text{eos})/\tau)}.
\end{align}
We apply Gaussian sampling only to global features, since patch-level features are instance-dependent and their patch-attribute correspondences are difficult to preserve under Gaussian sampling.
\begin{table*}[t]
	\vspace{-7mm}
	\caption{Comparison of the average and last performance of different methods. The best results are highlighted in bold. Detailed complete results are reported in the Appendix.}\label{tab:benchmark}
	\vspace{-2mm}
	\centering
	\resizebox{\textwidth}{!}{%
	\begin{NiceTabular}{@{} l *{15}{c}}
			\toprule
			\multicolumn{1}{c}{\multirow{3}{*}{Method}}
			&
			\multicolumn{4}{c}{Aircraft }   & 
			\multicolumn{4}{c}{CIFAR100 }	&	
			\multicolumn{4}{c}{Cars }   
			\\ 
			& 
			\multicolumn{2}{c}{B0 Inc10}   & 
			\multicolumn{2}{c}{B50 Inc10}	&		
			\multicolumn{2}{c}{B0 Inc10}   & 
			\multicolumn{2}{c}{B50 Inc10}	& 
			\multicolumn{2}{c}{B0 Inc10}   & 
			\multicolumn{2}{c}{B50 Inc10}	& 
			\\  
			& 
			{$\bar{\mathcal{A}}$} & ${\mathcal{A}_B}$  
			& {$\bar{\mathcal{A}}$} & ${\mathcal{A}_B}$
			& {$\bar{\mathcal{A}}$} & ${\mathcal{A}_B}$ 
			&  {$\bar{\mathcal{A}}$} & ${\mathcal{A}_B}$  
			& {$\bar{\mathcal{A}}$} & ${\mathcal{A}_B}$
			& {$\bar{\mathcal{A}}$} & ${\mathcal{A}_B}$ 
			\\
			\midrule
   ZS-CLIP~\cite{radford2021learning} &26.66 & 17.22 & 21.70 & 17.22& 81.81 & 71.38& 76.49 & 71.38& 82.60 & 76.37& 78.32 & 76.37\\
			\rowcolor{gray!10} SimpleCIL~\cite{zhou2025revisiting} &59.24 & 48.09 & 53.05 & 48.09 & 84.15 & 76.63& 80.20 & 76.63& 92.04 & 86.85 & 88.96 & 86.85\\
			L2P~\cite{wang2022learning}  &47.19 & 28.29 &44.07&32.13& 82.74 & 73.03& 81.14 & 73.61& 76.63 & 61.82& 76.37 & 65.64 \\
   \rowcolor{gray!10}
			DualPrompt~\cite{wang2022dualprompt}  & 44.30& 25.83 &46.07&33.57 & 81.63 & 72.44& 80.12 & 72.57& 76.26 & 62.94& 76.88 & 67.55 \\
			CODA-Prompt~\cite{smith2023coda}  & 45.98 & 27.69 & 45.14 & 32.28& 82.43 & 73.43& 78.69 & 71.58& 80.21 & 66.47& 75.06 & 64.19 \\
   \rowcolor{gray!10}
			RAPF~\cite{huang2024class}   &  50.38  & 23.61 &  40.47 &  25.44 & 86.14 & 78.04 & 82.17 &  77.93  & 82.89 & 62.85 &  75.87 & 63.19\\
   CLG-CBM~\citep{yu2025language}  & 66.05   &55.93 &59.25 & 55.39 &  86.58 & 80.15  &83.59  &79.28 &93.25  & 88.76    & 90.11  &88.19 \\
   \rowcolor{gray!10}
   PROOF~\citep{zhou2025learning}  &  63.81  & 56.14 &59.47 & 57.10& 86.77  &79.11   &83.32  & 79.73 &90.74  & 86.51    & 88.00  & 85.58\\
   BOFA~\citep{li2026bofa}  &70.96    &60.43  &66.09 &61.36 &  86.07 &  79.19 &  83.02&  79.44& 94.21 &  90.20 & 92.13  &90.50 \\
   \hline
	\rowcolor{LightCyan}\name (Ours) &\bf 71.57 & \bf61.51 & \bf66.82 & \bf 63.01 & \bf88.53 & \bf 81.81 &  \bf85.01 & \bf 81.60 &  \bf  94.43& \bf 90.91 & \bf 92.33 &  \bf 91.43 \\
        \end{NiceTabular}
	}	
	\resizebox{\textwidth}{!}{
        \begin{NiceTabular}{@{} l *{15}{c}}
			\toprule
			\multicolumn{1}{c}{\multirow{3}{*}{Method}}
			& 
			\multicolumn{4}{c}{ImageNet-R }   & 
			\multicolumn{4}{c}{CUB }	&	\multicolumn{4}{c}{UCF }   
			\\ 
			& 
			\multicolumn{2}{c}{B0 Inc20}   & 
			\multicolumn{2}{c}{B100 Inc20}	&	\multicolumn{2}{c}{B0 Inc20}   & 
			\multicolumn{2}{c}{B100 Inc20}	& 
			\multicolumn{2}{c}{B0 Inc10}   & 
			\multicolumn{2}{c}{B50 Inc10}	& 
			\\  
			& 
			{$\bar{\mathcal{A}}$} & ${\mathcal{A}_B}$  
			& {$\bar{\mathcal{A}}$} & ${\mathcal{A}_B}$
			& {$\bar{\mathcal{A}}$} & ${\mathcal{A}_B}$ 
			&  {$\bar{\mathcal{A}}$} & ${\mathcal{A}_B}$  
			& {$\bar{\mathcal{A}}$} & ${\mathcal{A}_B}$
			& {$\bar{\mathcal{A}}$} & ${\mathcal{A}_B}$ 
			\\
			\midrule
			ZS-CLIP~\cite{radford2021learning} &83.37 & 77.17& 79.57 & 77.17 & 74.38 & 63.06& 67.96 & 63.06& 75.50 & 67.64& 71.44 & 67.64\\
    \rowcolor{gray!10}
			SimpleCIL~\cite{zhou2025revisiting} & 81.06 & 74.48& 76.84 & 74.48& 83.81 & 77.52& 79.75 & 77.52& 90.44 & 85.68& 88.12 & 85.68\\
			L2P~\cite{wang2022learning}  &75.97 & 66.52 & 72.82 & 66.77&   70.87&57.93 & 75.64 &66.12 & 86.34 & 76.43& 83.95 & 76.62 \\
    \rowcolor{gray!10}
			DualPrompt~\cite{wang2022dualprompt}  &76.21 & 66.65 & 73.22 & 67.58&69.89 &57.46 & 74.40 &64.84 & 85.21 & 75.82& 84.31 & 76.35 \\
			CODA-Prompt~\cite{smith2023coda}  & 77.69 & 68.95 & 73.71 & 68.05& 73.12&62.98 &73.95&62.21 & 87.76 & 80.14&83.04 & 75.03 \\
    \rowcolor{gray!10}
			RAPF~\cite{huang2024class}  & 81.26  & 70.48 & 76.10 & 70.23 &  79.09 & 62.77& 72.82 & 62.93 & 92.28 & 80.33&90.31 & 81.55\\
             CLG-CBM~\citep{yu2025language}  &  84.64  & 78.50& 81.46& 77.88 &  85.37 & 78.24  &77.74  &76.97 & 95.04&  91.36   & 94.17  &91.85 \\
     \rowcolor{gray!10}
   PROOF~\citep{zhou2025learning}  &83.84    &78.40  &81.20 & 78.92&82.31   & 76.64  &79.20  &76.37  & 94.58 & 91.10    &  93.58 &90.91 \\
   BOFA~\citep{li2026bofa}   & 84.53   &78.77  & 81.60 &79.12 & 86.66  &80.58   & 83.18 & 80.79 & 93.19 &  88.71  &92.60   &89.43 \\
 \hline
  \rowcolor{LightCyan}\name (Ours)  & \bf 85.63 & \bf 79.08 & \bf 82.50 & \bf 79.50 & \bf87.17 & \bf81.93 &  \bf 84.43& \bf 82.23&  \bf 95.63 & \bf 92.38  & \bf 95.43 &  \bf 93.48  \\
			\bottomrule
		\end{NiceTabular}
	}
    \vspace{-4mm}
\end{table*}

\textbf{Summary.} In \mame, we exploit the local representations in CLIP through semantic-guided selection and structured alignment. Class-wise semantics are first constructed to select discriminative patches, and optimal transport is then used to establish correspondences between these patches and textual attributes. To adapt to incremental tasks and mitigate forgetting, task-specific projectors are introduced for visual and textual features. During training, we jointly optimize the global loss  and the local loss:
\begin{align}\label{eq:total-loss}
\mathcal{L} = \mathcal{L}_{g} + \beta\mathcal{L}_{l},
\end{align}
where $\beta$ is a hyperparameter balancing the two loss terms.\\
\textbf{Inference.} During inference, 
we aggregate the global and local logits to obtain the final prediction:
\begin{align}
    f(\mathbf{x}) = f_{\text{global}}(\mathbf{x}) + \beta f_{\text{local}}(\mathbf{x}).
\end{align}

\section{Experiments} \label{experiments}
In this section, we evaluate \name on nine benchmark datasets and compare it with state-of-the-art 
CIL methods. We then conduct ablation studies and additional analyses to examine the contribution of each component and the reliability of \mame. More experimental results are provided in the appendix.
\begin{figure*} 
\vspace{-7mm}
	\centering
	\begin{subfigure}{0.32\textwidth}		\includegraphics[width=0.95\columnwidth]{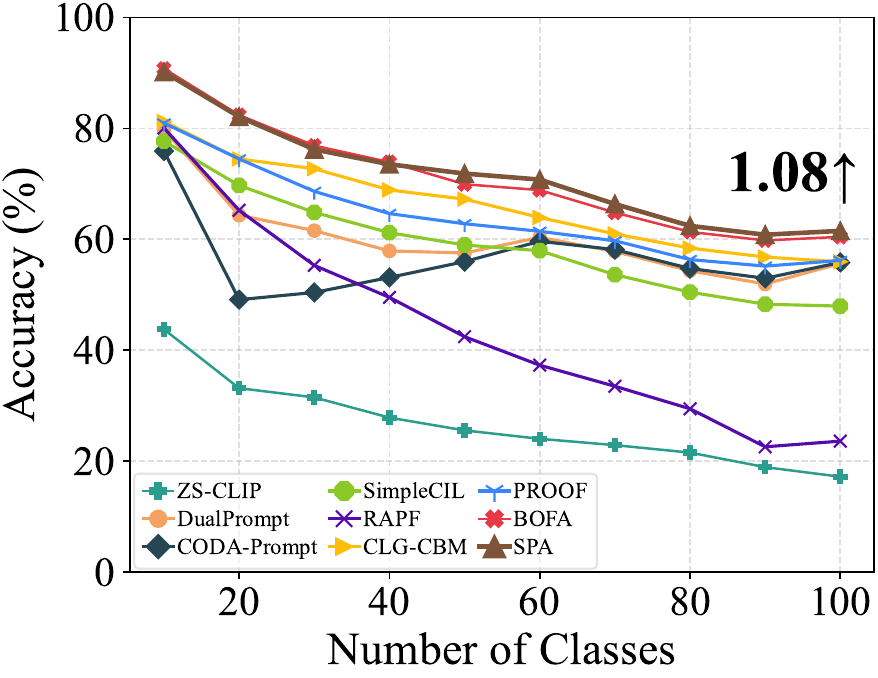}
		\caption{Aircraft Base0 Inc10}
	\end{subfigure}
	\hfill
	\begin{subfigure}{0.32\linewidth}
		\includegraphics[width=0.95\linewidth]{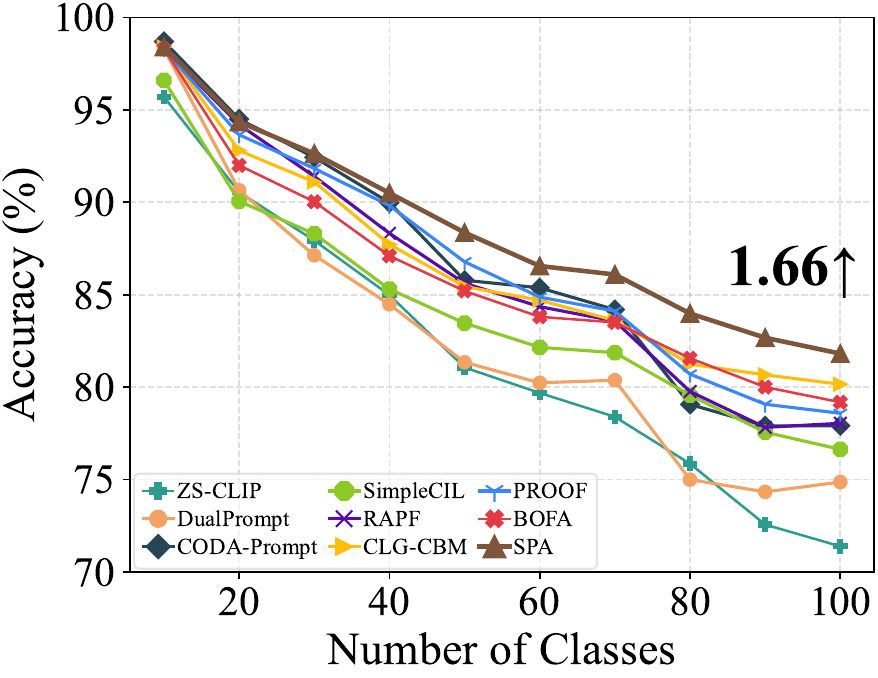}
		\caption{CIFAR100 Base0 Inc10}
	\end{subfigure}
	\hfill
	\begin{subfigure}{0.32\linewidth}
		\includegraphics[width=0.95\linewidth]{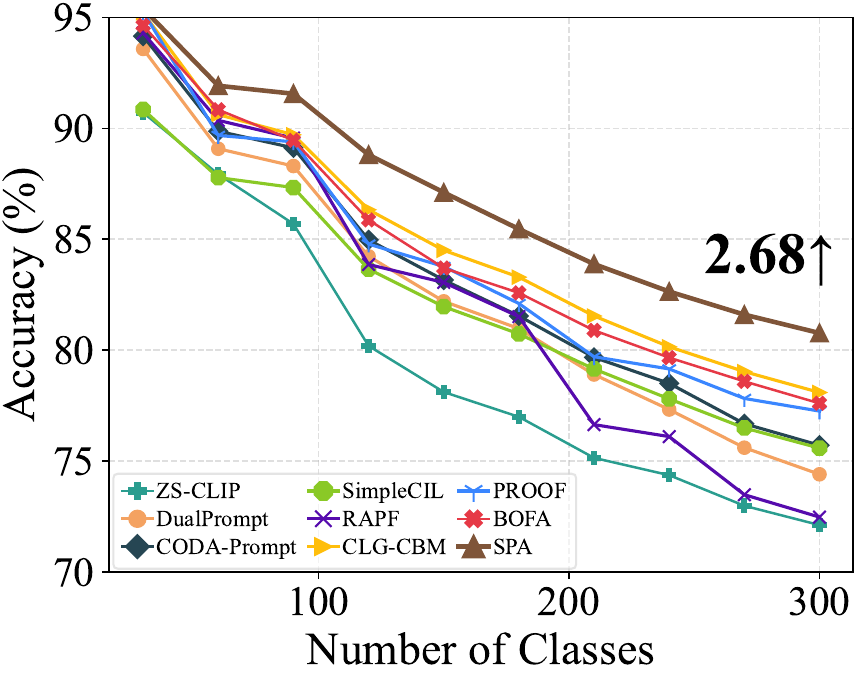}
		\caption{SUN Base0 Inc30}
	\end{subfigure}
	\\
	
	\vspace{-1mm}
	\caption{Incremental performance of different methods on the B0 setting. We report the performance gap after the last incremental stage of \name and the runner-up method at the end of the line.}
		\vspace{-5mm}
	\label{fig:supp-benchmark-b0}
\end{figure*}
\subsection{Implementation Details}
\textbf{Dataset.}  Following~\citep{zhou2025learning,zhou2022learning}, we evaluate our method on nine benchmark datasets, \textit{i.e.}, {CIFAR100}~\citep{krizhevsky2009learning}, {FGVCAircraft}~\citep{maji2013fine}, {CUB200}~\citep{wah2011caltech}, {ObjectNet}~\citep{barbu2019objectnet}, {Food101}~\citep{bossard2014food}, {ImageNet-R}~\cite{hendrycks2021many}, {StanfordCars}~\citep{krause20133d}, {UCF101}~\citep{soomro2012ucf101} and {SUN397}~\citep{xiao2010sun}. Adopting the sampling strategy from~\citep{zhou2025external,zhou2025learning}, we select 100 classes from CIFAR100, Food101, FGVCAircraft, StanfordCars, and UCF101; 200 classes from ObjectNet, CUB200, and ImageNet-R; and 300 classes from SUN397.\\
\textbf{Dataset split.} We construct the CIL tasks under the widely adopted `B-$m$ Inc-$n$' protocol~\citep{rebuffi2017icarl,wang2022learning}, where $m$ and $n$ denote the number of classes in the initial session and each subsequent incremental session, respectively. For fair comparison, the class order is randomly shuffled with a fixed random seed of 1993~\cite{rebuffi2017icarl} and kept consistent across all evaluated baselines.\\
\textbf{Comparison methods.}
To thoroughly evaluate the effectiveness of our proposed method, we compare it with SOTA CIL baselines. Specifically, these baselines include ViT-based methods, \textit{e.g.} SimpleCIL~\cite{zhou2025revisiting}, L2P~\citep{wang2022learning}, DualPrompt~\citep{wang2022dualprompt}, and CODA-Prompt~\citep{smith2023coda}, as well as CLIP-based methods, \textit{e.g.} RAPF~\citep{huang2024class}, CLG-CBM~\citep{yu2025language}, PROOF~\citep{zhou2025learning}, and BOFA~\citep{li2026bofa}. In addition, we use ZS-CLIP~\citep{radford2021learning} as a performance reference for CLIP on downstream tasks. More details are provided in the appendix.
\\\noindent \textbf{Training details.} All experiments are implemented in PyTorch~\citep{paszke2019pytorch} and conducted based on the C3Box~\citep{sun2026c3box} toolbox. Following~\citep{zhou2025external,zhou2025learning}, for fair comparison, \textbf{all compared methods use the same pre-trained CLIP backbone}, \textit{i.e.}, ViT-B/16. For visual-only methods (\textit{e.g.}, L2P, DualPrompt, CODA-Prompt) that cannot exploit textual priors, we initialize them with the visual branch of the same pre-trained CLIP backbone. We present the results based on LAION-400M pre-trained CLIP~\citep{ilharco2021openclip} in the main paper, while the corresponding results based on OpenAI pre-trained CLIP~\citep{radford2021learning} are provided in the appendix. We train our method using SGD for 10 epochs with a batch size of 64 and an initial learning rate of 0.05, which is decayed using a cosine annealing schedule. By default, the number of selected patches $K$ is set to 8, and the loss balance factor $\beta$ is set to 0.2. In addition, we use GPT-5.4~\citep{achiam2023gpt} to generate class-wise attribute semantics, and randomly sample $N = 5$ attribute descriptions for each class. Results with other LLMs are provided in the appendix.\\
\textbf{Evaluation metric.} Following the common CIL protocol~\citep{rebuffi2017icarl, zhou2025learning}, we evaluate all methods using the last accuracy $\mathcal{A}_B$ and the average accuracy $\bar{\mathcal{A}}=\frac{1}{B}\sum_{b=1}^{B}\mathcal{A}_b$, where $\mathcal{A}_b$ denotes the test accuracy 
after the $b$-th incremental session. Specifically, $\mathcal{A}_B$ denotes the accuracy over all learned classes after the final session, reflecting the model’s final performance. $\bar{\mathcal{A}}$ denotes the average accuracy across all sessions, measuring the overall performance and stability during incremental learning.
\subsection{Benchmark Comparison} \label{sec:5.2}
We first compare  \name with SOTA CIL methods on nine benchmark datasets, and report the results in Table~\ref{tab:benchmark} and Fig.~\ref{fig:supp-benchmark-b0}. Overall, \name consistently achieves the best performance, demonstrating the effectiveness of patch-level alignment for CLIP-based CIL. Visual prompt-based methods such as L2P and CODA-Prompt perform less competitively, as they cannot exploit rich textual semantics. CLIP-based methods such as RAPF and CLG-CBM achieve better performance by leveraging CLIP’s cross-modal representation capability. However, these methods still mainly rely on global alignment. In contrast, \name introduces semantic-guided patch selection and structured patch-level alignment, thereby learning more discriminative representations 
and mitigating catastrophic forgetting.

\subsection{Further Analysis}
\begin{figure*}[t]
  \vspace{-7mm}
 \centering
    \begin{subfigure}[t]{0.32\textwidth}
        \centering
        \includegraphics[width=0.95\linewidth]{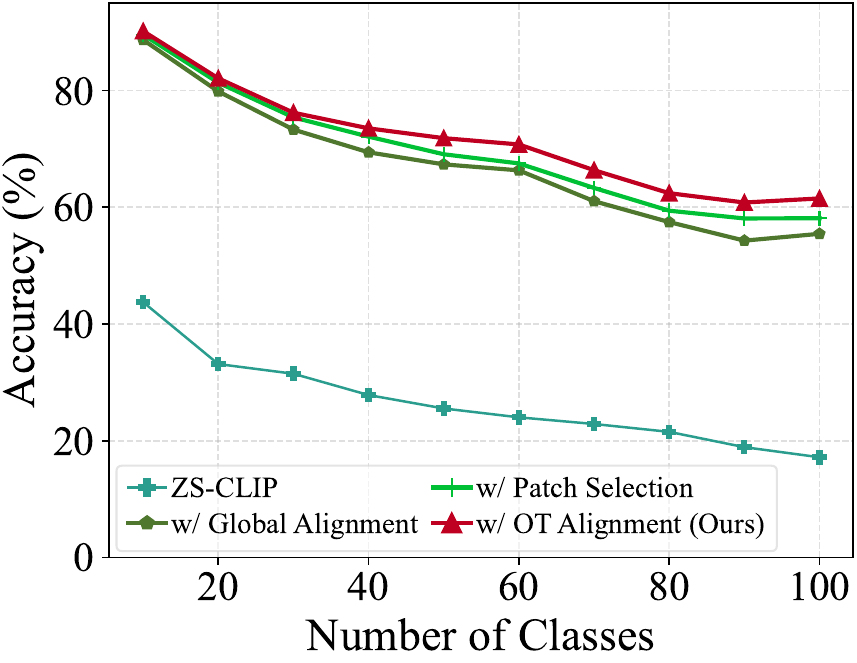}
        \caption{Ablation study}
        \label{fig:ablation}
    \end{subfigure}
    \hfill
   \begin{subfigure}[t]{0.32\textwidth}
        \centering
        \includegraphics[width=0.95\linewidth]{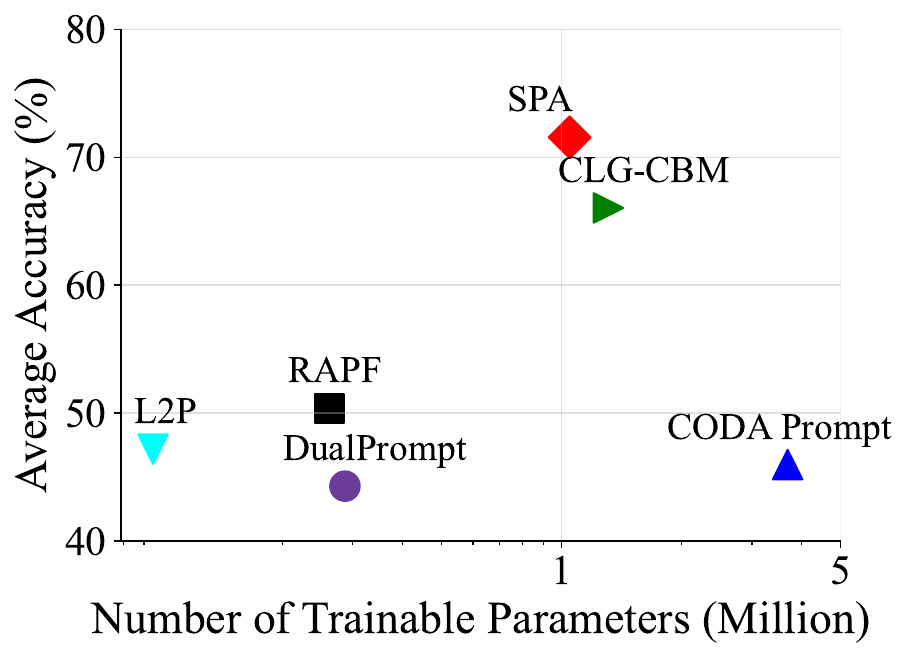}
        \caption{Trainable parameters}
        \label{fig:trainable}
    \end{subfigure}
     \begin{subfigure}[t]{0.32\textwidth}
        \centering
        \includegraphics[width=0.95\linewidth]{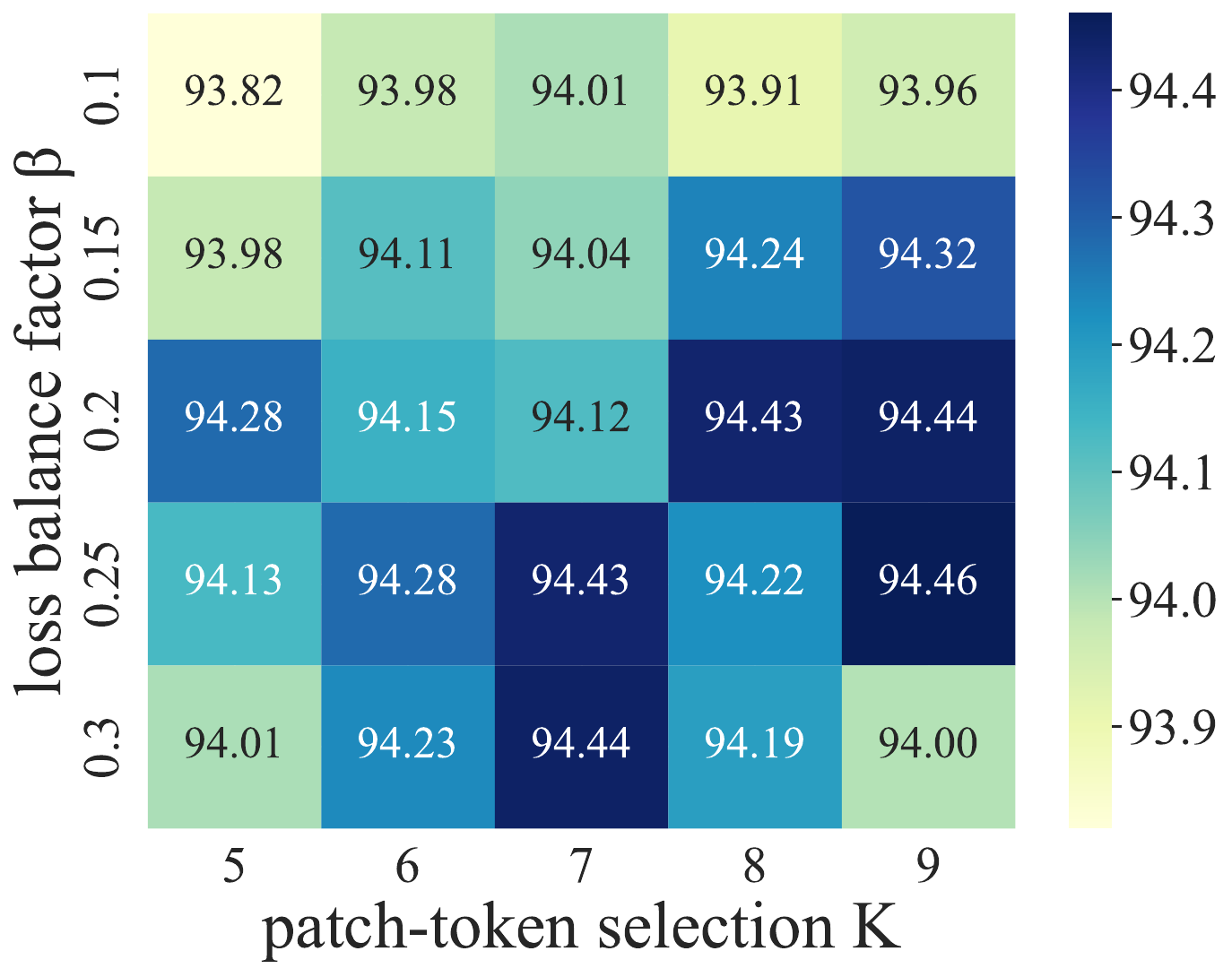}
        \caption{Parameter sensitivity}
        \label{fig:sensitivity}
    \end{subfigure}
    \vspace{-1mm}
    \caption{Ablation study, trainable parameters, and parameter sensitivity.}
    \label{fig:all-experiments}
\end{figure*}
\textbf{Ablation study.} We conduct ablation studies to analyze the contribution of each component in \name on Aircraft B0 Inc10, and report the results in Fig.~\ref{fig:ablation}. Specifically, we take \textbf{``ZS-CLIP''} as the baseline, whose performance is relatively limited. Based on this, we first introduce global alignment (Eq.~\ref{eq:unit-loss}), denoted as \textbf{``w/ Global Alignment''}, which brings clear performance gains and demonstrates the effectiveness of adapting CLIP to downstream incremental learning. We further introduce discriminative patch selection, denoted as \textbf{``w/ Patch Selection''}, which selects discriminative patches under the guidance of class-wise semantics using Eq.~\ref{eq:dis_set}, and aligns them with these semantics via a simple matching strategy. The further performance gain shows that semantic-guided patch selection reduces irrelevant visual information and improves the effectiveness of local representations in classification. Finally, we introduce optimal transport-based structured alignment (Eq.~\ref{eq:local-loss}), denoted as \textbf{``w/ OT Alignment''}, which corresponds to the full  \name and achieves the best performance. This shows that balanced matching further improves CIL performance. \\
\textbf{Trainable parameters.}
 We compare the trainable parameters and average accuracy on Aircraft B0 Inc10 in Fig.~\ref{fig:trainable}. The trainable parameters of \name mainly come from the lightweight task-specific projectors. Compared with prompt-based methods, \textit{e.g.}, L2P, \name introduces more parameters but achieves higher accuracy. Compared with CLIP-based methods, \name achieves the best performance with  a moderate parameter count, showing a favorable accuracy-parameter trade-off.\\
\noindent\textbf{Parameter robustness.} We analyze the sensitivity of our method on Cars B0 Inc10 by varying two key hyperparameters: the number of selected patches $K$ and the loss balance factor $\beta$. Specifically, we set $K \in \{5,6,7,8,9\}$ and $\beta \in \{0.1,0.15,0.2,0.25,0.3\}$. We report the average  performance in Fig.~\ref{fig:sensitivity}. As shown in the figure, the performance remains relatively stable across different choices of $K$ and $\beta$. A proper patch number can provide a stable trade-off between preserving informative local information and suppressing redundant noise.\\
\noindent\textbf{Visualizations.} 
\begin{figure*}
	\vspace{-4mm}
	\centering
	       \begin{minipage}{\textwidth}
        \centering
        \includegraphics[width=0.95\textwidth]{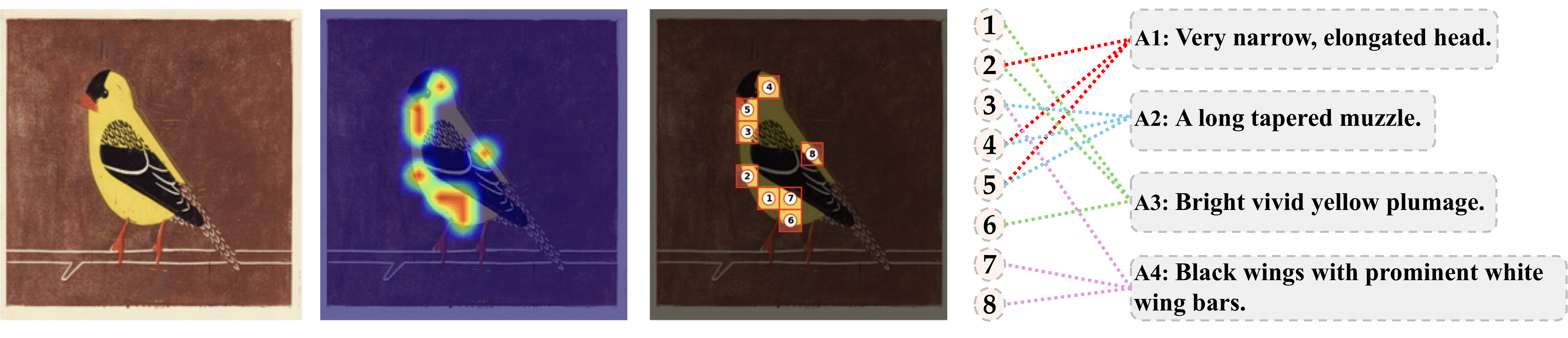} \\
        \label{fig:input}
    \end{minipage}
 \vspace{-2mm}
	\caption{ {\bf Column 1:} Input.  {\bf Column 2:} Response heatmap of discriminative patches.  {\bf Column 3:} Visualization of the top-$8$ discriminative patches selected. {\bf Column 4:} OT-based correspondence graph, where numbered nodes denote the selected patches and attribute nodes denote class-wise attributes. For clarity, only the top-3 correspondences with the highest transport weights are shown.}
	\vspace{-6mm}
	\label{fig:vis}
\end{figure*}
Fig.~\ref{fig:vis} visualizes the proposed OT-based patch-level alignment. The selection heatmaps (\textbf{Column 2}) show that class-wise attributes guide the model to focus on discriminative local regions. The OT-based correspondence graphs (\textbf{Column 4}) further reveal structured associations between selected patches and class-wise semantic attributes. For example, the attribute \textit{``very narrow, elongated head''} describes the bird head and is assigned a strong correspondence to patch \textit{4} on the head region. These results show that \name can build meaningful cross-modal associations between local visual regions and semantic attributes. More visualization results are provided in the Appendix.

\section{Conclusion}\label{conclusion}
We propose \mame, a semantic-guided patch-level alignment framework for CLIP-based CIL. Different from previous methods that mainly rely on global alignment and overlook patch-level information, \name explicitly exploits the rich local representations in CLIP. \name first constructs diverse visual samples for each class, and then leverages GPT to generate class-wise semantics. These semantics are used to select discriminative patches and align them with textual attributes through optimal transport, while task-specific projectors are introduced to mitigate catastrophic forgetting. Extensive experiments demonstrate that \name achieves strong performance across multiple benchmarks.\\
\noindent\textbf{Limitations and Future Work.}
\name still depends on externally generated attribute semantics, whose quality may affect patch selection and local alignment. In addition, optimal transport introduces extra computation. Future work will explore more efficient and robust patch-level alignment strategies.

{
    \small
    \bibliographystyle{ieeenat_fullname}
    \bibliography{main}

@inproceedings{hendrycks2021many,
  title={The many faces of robustness: A critical analysis of out-of-distribution generalization},
  author={Hendrycks, Dan and Basart, Steven and Mu, Norman and Kadavath, Saurav and Wang, Frank and Dorundo, Evan and Desai, Rahul and Zhu, Tyler and Parajuli, Samyak and Guo, Mike and others},
  booktitle={Proceedings of the IEEE/CVF international conference on computer vision},
  pages={8340--8349},
  year={2021}
}

@inproceedings{he2016deep,
  title={Deep residual learning for image recognition},
  author={He, Kaiming and Zhang, Xiangyu and Ren, Shaoqing and Sun, Jian},
  booktitle={Proceedings of the IEEE conference on computer vision and pattern recognition},
  pages={770--778},
  year={2016}
}

@inproceedings{liu2015deep,
  title={Deep learning face attributes in the wild},
  author={Liu, Ziwei and Luo, Ping and Wang, Xiaogang and Tang, Xiaoou},
  booktitle={Proceedings of the IEEE international conference on computer vision},
  pages={3730--3738},
  year={2015}
}

@inproceedings{rebuffi2017icarl,
  title={icarl: Incremental classifier and representation learning},
  author={Rebuffi, Sylvestre-Alvise and Kolesnikov, Alexander and Sperl, Georg and Lampert, Christoph H},
  booktitle={Proceedings of the IEEE conference on Computer Vision and Pattern Recognition},
  pages={2001--2010},
  year={2017}
}

@inproceedings{zhao2020maintaining,
  title={Maintaining discrimination and fairness in class incremental learning},
  author={Zhao, Bowen and Xiao, Xi and Gan, Guojun and Zhang, Bin and Xia, Shu-Tao},
  booktitle={Proceedings of the IEEE/CVF conference on computer vision and pattern recognition},
  pages={13208--13217},
  year={2020}
}

@article{shi2021overcoming,
  title={Overcoming catastrophic forgetting in incremental few-shot learning by finding flat minima},
  author={Shi, Guangyuan and Chen, Jiaxin and Zhang, Wenlong and Zhan, Li-Ming and Wu, Xiao-Ming},
  journal={Advances in neural information processing systems},
  volume={34},
  pages={6747--6761},
  year={2021}
}

@inproceedings{serra2018overcoming,
  title={Overcoming catastrophic forgetting with hard attention to the task},
  author={Serra, Joan and Suris, Didac and Miron, Marius and Karatzoglou, Alexandros},
  booktitle={International conference on machine learning},
  pages={4548--4557},
  year={2018},
  organization={PMLR}
}

@article{de2021continual,
  title={A continual learning survey: Defying forgetting in classification tasks},
  author={De Lange, Matthias and Aljundi, Rahaf and Masana, Marc and Parisot, Sarah and Jia, Xu and Leonardis, Ale{\v{s}} and Slabaugh, Gregory and Tuytelaars, Tinne},
  journal={IEEE transactions on pattern analysis and machine intelligence},
  volume={44},
  number={7},
  pages={3366--3385},
  year={2021},
  publisher={IEEE}
}

@article{zhou2025revisiting,
  title={Revisiting class-incremental learning with pre-trained models: Generalizability and adaptivity are all you need},
  author={Zhou, Da-Wei and Cai, Zi-Wen and Ye, Han-Jia and Zhan, De-Chuan and Liu, Ziwei},
  journal={International Journal of Computer Vision},
  volume={133},
  number={3},
  pages={1012--1032},
  year={2025},
  publisher={Springer}
}

@inproceedings{gao2022r,
  title={R-dfcil: Relation-guided representation learning for data-free class incremental learning},
  author={Gao, Qiankun and Zhao, Chen and Ghanem, Bernard and Zhang, Jian},
  booktitle={European Conference on Computer Vision},
  pages={423--439},
  year={2022},
  organization={Springer}
}

@article{french1999catastrophic,
  title={Catastrophic forgetting in connectionist networks},
  author={French, Robert M},
  journal={Trends in cognitive sciences},
  volume={3},
  number={4},
  pages={128--135},
  year={1999},
  publisher={Elsevier}
}

@article{zhou2024class,
  title={Class-incremental learning: A survey},
  author={Zhou, Da-Wei and Wang, Qi-Wei and Qi, Zhi-Hong and Ye, Han-Jia and Zhan, De-Chuan and Liu, Ziwei},
  journal={IEEE Transactions on Pattern Analysis and Machine Intelligence},
  volume={46},
  number={12},
  pages={9851--9873},
  year={2024},
  publisher={IEEE}
}

@article{bai2023qwen,
  title={Qwen-vl: A versatile vision-language model for understanding, localization},
  author={Bai, Jinze and Bai, Shuai and Yang, Shusheng and Wang, Shijie and Tan, Sinan and Wang, Peng and Lin, Junyang and Zhou, Chang and Zhou, Jingren},
  journal={Text Reading, and Beyond},
  volume={2},
  number={1},
  pages={1},
  year={2023}
}

@inproceedings{zhou2024continual,
  title={Continual learning with pre-trained models: a survey},
  author={Zhou, Da-Wei and Sun, Hai-Long and Ning, Jingyi and Ye, Han-Jia and Zhan, De-Chuan},
  booktitle={Proceedings of the Thirty-Third International Joint Conference on Artificial Intelligence},
  pages={8363--8371},
  year={2024}
}

@inproceedings{wang2022learning,
  title={Learning to prompt for continual learning},
  author={Wang, Zifeng and Zhang, Zizhao and Lee, Chen-Yu and Zhang, Han and Sun, Ruoxi and Ren, Xiaoqi and Su, Guolong and Perot, Vincent and Dy, Jennifer and Pfister, Tomas},
  booktitle={Proceedings of the IEEE/CVF conference on computer vision and pattern recognition},
  pages={139--149},
  year={2022}
}

@inproceedings{li2023blip,
  title={Blip-2: Bootstrapping language-image pre-training with frozen image encoders and large language models},
  author={Li, Junnan and Li, Dongxu and Savarese, Silvio and Hoi, Steven},
  booktitle={International conference on machine learning},
  pages={19730--19742},
  year={2023},
  organization={PMLR}
}

@inproceedings{yao2022pevl,
  title={PEVL: Position-enhanced pre-training and prompt tuning for vision-language models},
  author={Yao, Yuan and Chen, Qianyu and Zhang, Ao and Ji, Wei and Liu, Zhiyuan and Chua, Tat-Seng and Sun, Maosong},
  booktitle={Proceedings of the 2022 conference on empirical methods in natural language processing},
  pages={11104--11117},
  year={2022}
}

@inproceedings{radford2021learning,
  title={Learning transferable visual models from natural language supervision},
  author={Radford, Alec and Kim, Jong Wook and Hallacy, Chris and Ramesh, Aditya and Goh, Gabriel and Agarwal, Sandhini and Sastry, Girish and Askell, Amanda and Mishkin, Pamela and Clark, Jack and others},
  booktitle={International conference on machine learning},
  pages={8748--8763},
  year={2021},
  organization={PmLR}
}

@article{dosovitskiy2020image,
  title={An image is worth 16x16 words: Transformers for image recognition at scale},
  author={Dosovitskiy, Alexey and Beyer, Lucas and Kolesnikov, Alexander and Weissenborn, Dirk and Zhai, Xiaohua and Unterthiner, Thomas and Dehghani, Mostafa and Minderer, Matthias and Heigold, Georg and Gelly, Sylvain and others},
  journal={arXiv preprint arXiv:2010.11929},
  year={2020}
}

@inproceedings{jia2021scaling,
  title={Scaling up visual and vision-language representation learning with noisy text supervision},
  author={Jia, Chao and Yang, Yinfei and Xia, Ye and Chen, Yi-Ting and Parekh, Zarana and Pham, Hieu and Le, Quoc and Sung, Yun-Hsuan and Li, Zhen and Duerig, Tom},
  booktitle={International conference on machine learning},
  pages={4904--4916},
  year={2021},
  organization={PMLR}
}

@article{zhou2025learning,
  title={Learning without forgetting for vision-language models},
  author={Zhou, Da-Wei and Zhang, Yuanhan and Wang, Yan and Ning, Jingyi and Ye, Han-Jia and Zhan, De-Chuan and Liu, Ziwei},
  journal={IEEE Transactions on Pattern Analysis and Machine Intelligence},
  year={2025},
  publisher={IEEE}
}

@inproceedings{huang2025mind,
  title={Mind the gap: Preserving and compensating for the modality gap in clip-based continual learning},
  author={Huang, Linlan and Cao, Xusheng and Lu, Haori and Meng, Yifan and Yang, Fei and Liu, Xialei},
  booktitle={Proceedings of the IEEE/CVF International Conference on Computer Vision},
  pages={3777--3786},
  year={2025}
}

@article{hu2025hierarchical,
  title={Hierarchical Semantic Tree Anchoring for CLIP-Based Class-Incremental Learning},
  author={Hu, Tao and Li, Lan and Xie, Zhen-Hao and Zhou, Da-Wei},
  journal={arXiv preprint arXiv:2511.15633},
  year={2025}
}

@inproceedings{wu2024llm2clip,
  title={Llm2clip: Powerful language model unlock richer visual representation},
  author={Wu, Aoqi and Yang, Yifan and Luo, Xufang and Yang, Yuqing and Wang, Chunyu and Hu, Liang and Dai, Xiyang and Chen, Dongdong and Luo, Chong and Qiu, Lili and others},
  booktitle={NeurIPS 2024 Workshop: Self-Supervised Learning-Theory and Practice},
  year={2024}
}

@inproceedings{zhou2022conditional,
  title={Conditional prompt learning for vision-language models},
  author={Zhou, Kaiyang and Yang, Jingkang and Loy, Chen Change and Liu, Ziwei},
  booktitle={Proceedings of the IEEE/CVF conference on computer vision and pattern recognition},
  pages={16816--16825},
  year={2022}
}

@article{wu2024controlmllm,
  title={Controlmllm: Training-free visual prompt learning for multimodal large language models},
  author={Wu, Mingrui and Cai, Xinyue and Ji, Jiayi and Li, Jiale and Huang, Oucheng and Fei, Hao and Jiang, Guannan and Sun, Xiaoshuai and Ji, Rongrong},
  journal={Advances in Neural Information Processing Systems},
  volume={37},
  pages={45206--45234},
  year={2024}
}

@inproceedings{yu2024boosting,
  title={Boosting continual learning of vision-language models via mixture-of-experts adapters},
  author={Yu, Jiazuo and Zhuge, Yunzhi and Zhang, Lu and Hu, Ping and Wang, Dong and Lu, Huchuan and He, You},
  booktitle={Proceedings of the IEEE/CVF Conference on Computer Vision and Pattern Recognition},
  pages={23219--23230},
  year={2024}
}

@article{gao2024clip,
  title={Clip-adapter: Better vision-language models with feature adapters},
  author={Gao, Peng and Geng, Shijie and Zhang, Renrui and Ma, Teli and Fang, Rongyao and Zhang, Yongfeng and Li, Hongsheng and Qiao, Yu},
  journal={International journal of computer vision},
  volume={132},
  number={2},
  pages={581--595},
  year={2024},
  publisher={Springer}
}

@article{chen2022plot,
  title={Plot: Prompt learning with optimal transport for vision-language models},
  author={Chen, Guangyi and Yao, Weiran and Song, Xiangchen and Li, Xinyue and Rao, Yongming and Zhang, Kun},
  journal={arXiv preprint arXiv:2210.01253},
  year={2022}
}

@inproceedings{lafon2024gallop,
  title={Gallop: Learning global and local prompts for vision-language models},
  author={Lafon, Marc and Ramzi, Elias and Rambour, Cl{\'e}ment and Audebert, Nicolas and Thome, Nicolas},
  booktitle={European Conference on Computer Vision},
  pages={264--282},
  year={2024},
  organization={Springer}
}

@article{masana2022class,
  title={Class-incremental learning: survey and performance evaluation on image classification},
  author={Masana, Marc and Liu, Xialei and Twardowski, Bart{\l}omiej and Menta, Mikel and Bagdanov, Andrew D and Van De Weijer, Joost},
  journal={IEEE Transactions on Pattern Analysis and Machine Intelligence},
  volume={45},
  number={5},
  pages={5513--5533},
  year={2022},
  publisher={IEEE}
}

@inproceedings{zenke2017continual,
  title={Continual learning through synaptic intelligence},
  author={Zenke, Friedemann and Poole, Ben and Ganguli, Surya},
  booktitle={International conference on machine learning},
  pages={3987--3995},
  year={2017},
  organization={Pmlr}
}

@article{kirkpatrick2017overcoming,
  title={Overcoming catastrophic forgetting in neural networks},
  author={Kirkpatrick, James and Pascanu, Razvan and Rabinowitz, Neil and Veness, Joel and Desjardins, Guillaume and Rusu, Andrei A and Milan, Kieran and Quan, John and Ramalho, Tiago and Grabska-Barwinska, Agnieszka and others},
  journal={Proceedings of the national academy of sciences},
  volume={114},
  number={13},
  pages={3521--3526},
  year={2017},
  publisher={National Academy of Sciences}
}

@inproceedings{aljundi2018memory,
  title={Memory aware synapses: Learning what (not) to forget},
  author={Aljundi, Rahaf and Babiloni, Francesca and Elhoseiny, Mohamed and Rohrbach, Marcus and Tuytelaars, Tinne},
  booktitle={Proceedings of the European conference on computer vision (ECCV)},
  pages={139--154},
  year={2018}
}

@inproceedings{xiang2019incremental,
  title={Incremental learning using conditional adversarial networks},
  author={Xiang, Ye and Fu, Ying and Ji, Pan and Huang, Hua},
  booktitle={Proceedings of the IEEE/CVF international conference on computer vision},
  pages={6619--6628},
  year={2019}
}

@inproceedings{ostapenko2019learning,
  title={Learning to remember: A synaptic plasticity driven framework for continual learning},
  author={Ostapenko, Oleksiy and Puscas, Mihai and Klein, Tassilo and Jahnichen, Patrick and Nabi, Moin},
  booktitle={Proceedings of the IEEE/CVF conference on computer vision and pattern recognition},
  pages={11321--11329},
  year={2019}
}

@inproceedings{luo2023class,
  title={Class-incremental exemplar compression for class-incremental learning},
  author={Luo, Zilin and Liu, Yaoyao and Schiele, Bernt and Sun, Qianru},
  booktitle={Proceedings of the IEEE/CVF Conference on Computer Vision and Pattern Recognition},
  pages={11371--11380},
  year={2023}
}

@article{chaudhry2018efficient,
  title={Efficient lifelong learning with a-gem},
  author={Chaudhry, Arslan and Ranzato, Marc'Aurelio and Rohrbach, Marcus and Elhoseiny, Mohamed},
  journal={arXiv preprint arXiv:1812.00420},
  year={2018}
}

@article{xu2018reinforced,
  title={Reinforced continual learning},
  author={Xu, Ju and Zhu, Zhanxing},
  journal={Advances in neural information processing systems},
  volume={31},
  year={2018}
}

@article{yoon2017lifelong,
  title={Lifelong learning with dynamically expandable networks},
  author={Yoon, Jaehong and Yang, Eunho and Lee, Jeongtae and Hwang, Sung Ju},
  journal={arXiv preprint arXiv:1708.01547},
  year={2017}
}

@inproceedings{wang2022beef,
  title={Beef: Bi-compatible class-incremental learning via energy-based expansion and fusion},
  author={Wang, Fu-Yun and Zhou, Da-Wei and Liu, Liu and Ye, Han-Jia and Bian, Yatao and Zhan, De-Chuan and Zhao, Peilin},
  booktitle={The eleventh international conference on learning representations},
  year={2022}
}

@inproceedings{zheng2025task,
  title={Task-agnostic guided feature expansion for class-incremental learning},
  author={Zheng, Bowen and Zhou, Da-Wei and Ye, Han-Jia and Zhan, De-Chuan},
  booktitle={Proceedings of the Computer Vision and Pattern Recognition Conference},
  pages={10099--10109},
  year={2025}
}

@inproceedings{liu2021adaptive,
  title={Adaptive aggregation networks for class-incremental learning},
  author={Liu, Yaoyao and Schiele, Bernt and Sun, Qianru},
  booktitle={Proceedings of the IEEE/CVF conference on Computer Vision and Pattern Recognition},
  pages={2544--2553},
  year={2021}
}

@article{hinton2015distilling,
  title={Distilling the knowledge in a neural network},
  author={Hinton, Geoffrey and Vinyals, Oriol and Dean, Jeff},
  journal={arXiv preprint arXiv:1503.02531},
  year={2015}
}

@article{li2017learning,
  title={Learning without forgetting},
  author={Li, Zhizhong and Hoiem, Derek},
  journal={IEEE transactions on pattern analysis and machine intelligence},
  volume={40},
  number={12},
  pages={2935--2947},
  year={2017},
  publisher={IEEE}
}

@inproceedings{wang2022dualprompt,
  title={Dualprompt: Complementary prompting for rehearsal-free continual learning},
  author={Wang, Zifeng and Zhang, Zizhao and Ebrahimi, Sayna and Sun, Ruoxi and Zhang, Han and Lee, Chen-Yu and Ren, Xiaoqi and Su, Guolong and Perot, Vincent and Dy, Jennifer and others},
  booktitle={European conference on computer vision},
  pages={631--648},
  year={2022},
  organization={Springer}
}

@article{zhou2022learning,
  title={Learning to prompt for vision-language models},
  author={Zhou, Kaiyang and Yang, Jingkang and Loy, Chen Change and Liu, Ziwei},
  journal={International journal of computer vision},
  volume={130},
  number={9},
  pages={2337--2348},
  year={2022},
  publisher={Springer}
}

@inproceedings{zhu2021prototype,
  title={Prototype augmentation and self-supervision for incremental learning},
  author={Zhu, Fei and Zhang, Xu-Yao and Wang, Chuang and Yin, Fei and Liu, Cheng-Lin},
  booktitle={Proceedings of the IEEE/CVF conference on computer vision and pattern recognition},
  pages={5871--5880},
  year={2021}
}

@inproceedings{huang2024class,
  title={Class-incremental learning with clip: Adaptive representation adjustment and parameter fusion},
  author={Huang, Linlan and Cao, Xusheng and Lu, Haori and Liu, Xialei},
  booktitle={European Conference on Computer Vision},
  pages={214--231},
  year={2024},
  organization={Springer}
}

@inproceedings{fukuda2025adapter,
  title={Adapter merging with centroid prototype mapping for scalable class-incremental learning},
  author={Fukuda, Takuma and Kera, Hiroshi and Kawamoto, Kazuhiko},
  booktitle={Proceedings of the computer vision and pattern recognition conference},
  pages={4884--4893},
  year={2025}
}

@inproceedings{tan2024semantically,
  title={Semantically-shifted incremental adapter-tuning is a continual vitransformer},
  author={Tan, Yuwen and Zhou, Qinhao and Xiang, Xiang and Wang, Ke and Wu, Yuchuan and Li, Yongbin},
  booktitle={Proceedings of the IEEE/CVF Conference on Computer Vision and Pattern Recognition},
  pages={23252--23262},
  year={2024}
}

@inproceedings{zhou2025external,
  title={External knowledge injection for clip-based class-incremental learning},
  author={Zhou, Da-Wei and Li, Kai-Wen and Ning, Jingyi and Ye, Han-Jia and Zhang, Lijun and Zhan, De-Chuan},
  booktitle={Proceedings of the IEEE/CVF International Conference on Computer Vision},
  pages={3314--3325},
  year={2025}
}

@inproceedings{yu2025language,
  title={Language guided concept bottleneck models for interpretable continual learning},
  author={Yu, Lu and Han, Haoyu and Tao, Zhe and Yao, Hantao and Xu, Changsheng},
  booktitle={Proceedings of the Computer Vision and Pattern Recognition Conference},
  pages={14976--14986},
  year={2025}
}

@article{li2025addressing,
  title={Addressing imbalanced domain-incremental learning through dual-balance collaborative experts},
  author={Li, Lan and Zhou, Da-Wei and Ye, Han-Jia and Zhan, De-Chuan},
  journal={arXiv preprint arXiv:2507.07100},
  year={2025}
}

@article{qi2025adaptive,
  title={Adaptive adapter routing for long-tailed class-incremental learning},
  author={Qi, Zhi-Hong and Zhou, Da-Wei and Yao, Yiran and Ye, Han-Jia and Zhan, De-Chuan},
  journal={Machine Learning},
  volume={114},
  number={3},
  pages={68},
  year={2025},
  publisher={Springer}
}

@inproceedings{smith2023coda,
  title={Coda-prompt: Continual decomposed attention-based prompting for rehearsal-free continual learning},
  author={Smith, James Seale and Karlinsky, Leonid and Gutta, Vyshnavi and Cascante-Bonilla, Paola and Kim, Donghyun and Arbelle, Assaf and Panda, Rameswar and Feris, Rogerio and Kira, Zsolt},
  booktitle={Proceedings of the IEEE/CVF conference on computer vision and pattern recognition},
  pages={11909--11919},
  year={2023}
}

@article{wang2022s,
  title={S-prompts learning with pre-trained transformers: An occam’s razor for domain incremental learning},
  author={Wang, Yabin and Huang, Zhiwu and Hong, Xiaopeng},
  journal={Advances in Neural Information Processing Systems},
  volume={35},
  pages={5682--5695},
  year={2022}
}

@article{krizhevsky2009learning,
  title={Learning multiple layers of features from tiny images},
  author={Krizhevsky, Alex and Hinton, Geoffrey and others},
  year={2009},
  publisher={Toronto, ON, Canada}
}

@article{wah2011caltech,
  title={The caltech-ucsd birds-200-2011 dataset},
  author={Wah, Catherine and Branson, Steve and Welinder, Peter and Perona, Pietro and Belongie, Serge},
  year={2011}
}

@article{barbu2019objectnet,
  title={Objectnet: A large-scale bias-controlled dataset for pushing the limits of object recognition models},
  author={Barbu, Andrei and Mayo, David and Alverio, Julian and Luo, William and Wang, Christopher and Gutfreund, Dan and Tenenbaum, Josh and Katz, Boris},
  journal={Advances in neural information processing systems},
  volume={32},
  year={2019}
}

@article{maji2013fine,
  title={Fine-grained visual classification of aircraft},
  author={Maji, Subhransu and Rahtu, Esa and Kannala, Juho and Blaschko, Matthew and Vedaldi, Andrea},
  journal={arXiv preprint arXiv:1306.5151},
  year={2013}
}

@inproceedings{bossard2014food,
  title={Food-101--mining discriminative components with random forests},
  author={Bossard, Lukas and Guillaumin, Matthieu and Van Gool, Luc},
  booktitle={European conference on computer vision},
  pages={446--461},
  year={2014},
  organization={Springer}
}

@inproceedings{xiao2010sun,
  title={Sun database: Large-scale scene recognition from abbey to zoo},
  author={Xiao, Jianxiong and Hays, James and Ehinger, Krista A and Oliva, Aude and Torralba, Antonio},
  booktitle={2010 IEEE computer society conference on computer vision and pattern recognition},
  pages={3485--3492},
  year={2010},
  organization={IEEE}
}

@article{soomro2012ucf101,
  title={Ucf101: A dataset of 101 human actions classes from videos in the wild},
  author={Soomro, Khurram and Zamir, Amir Roshan and Shah, Mubarak},
  journal={arXiv preprint arXiv:1212.0402},
  year={2012}
}

@inproceedings{krause20133d,
  title={3d object representations for fine-grained categorization},
  author={Krause, Jonathan and Stark, Michael and Deng, Jia and Fei-Fei, Li},
  booktitle={Proceedings of the IEEE international conference on computer vision workshops},
  pages={554--561},
  year={2013}
}

@inproceedings{li2026bofa,
  title={Bofa: Bridge-layer orthogonal low-rank fusion for clip-based class-incremental learning},
  author={Li, Lan and Hu, Tao and Zhou, Da-Wei and Yang, Jia-Qi and Ye, Han-Jia and Zhan, De-Chuan},
  booktitle={Proceedings of the AAAI Conference on Artificial Intelligence},
  volume={40},
  number={27},
  pages={22967--22975},
  year={2026}
}

@article{paszke2019pytorch,
  title={Pytorch: An imperative style, high-performance deep learning library},
  author={Paszke, Adam and Gross, Sam and Massa, Francisco and Lerer, Adam and Bradbury, James and Chanan, Gregory and Killeen, Trevor and Lin, Zeming and Gimelshein, Natalia and Antiga, Luca and others},
  journal={Advances in neural information processing systems},
  volume={32},
  year={2019}
}

@article{sun2026c3box,
  title={C3Box: A CLIP-based Class-Incremental Learning Toolbox},
  author={Sun, Hao and Zhou, Da-Wei},
  journal={arXiv preprint arXiv:2601.20852},
  year={2026}
}

@article{achiam2023gpt,
  title={Openai gpt-5 system card},
  author={Singh, Aaditya and Fry, Adam and Perelman, Adam and Tart, Adam and Ganesh, Adi and El-Kishky, Ahmed and McLaughlin, Aidan and Low, Aiden and Ostrow, AJ and Ananthram, Akhila and others},
  journal={arXiv preprint arXiv:2601.03267},
  year={2025}
}

@article{gangbo1996geometry,
  title={The geometry of optimal transportation},
  author={Gangbo, Wilfrid and McCann, Robert J},
  year={1996}
}

@article{cuturi2013sinkhorn,
  title={Sinkhorn distances: Lightspeed computation of optimal transport},
  author={Cuturi, Marco},
  journal={Advances in neural information processing systems},
  volume={26},
  year={2013}
}

@article{ilharco2021openclip,
  title={Openclip},
  author={Ilharco, Gabriel and Wortsman, Mitchell and Carlini, Nicholas and Taori, Rohan and Dave, Achal and Shankar, Vaishaal and Namkoong, Hongseok and Miller, John and Hajishirzi, Hannaneh and Farhadi, Ali and others},
  journal={Zenodo},
  year={2021}
}
}
\newpage
\appendix
\renewcommand{\thesection}{\Alph{section}}

\setcounter{section}{0}
\section*{\centering Appendix}
In this supplementary material, we provide more details
about \mame, including more implementation details and
experimental results. The appendix is organized as follows:

\textbf{Appendix.~\ref{sec:a1}} provides additional experimental analyses, including running time comparison, backbone robustness, multiple random class orders, LLM variants, top-$K$ patch selection strategies, parameter robustness, and forgetting analysis.

\textbf{Appendix.~\ref{sec:visualization}} provides visualizations of our method, including discriminative patch selection, OT-based patch-level alignment, and examples of generated class-wise attribute semantics.

\textbf{Appendix.~\ref{sec:alg}} describes the training algorithm of our method.

\textbf{Appendix.~\ref{sec:a2}}  introduces the compared methods used in experiments.

\textbf{Appendix.~\ref{sec:full-results}} reports full results across datasets and incremental settings.

\textbf{Appendix.~\ref{sec:impact}} discusses the broader impacts of \mame.
\section{More Results} \label{sec:a1}
In this section, we provide additional experimental analyses to complement the main paper. We first compare the running time of different methods, evaluate \name under different CLIP pre-trained weights, and report results with multiple random class orders. We then analyze the influence of different LLMs, trainable parameters, and top-$K$ patch selection strategies. 
\begin{figure*}[ht]
 \centering
    
    \begin{subfigure}[t]{0.32\textwidth}
        \centering
        \includegraphics[width=1\linewidth]{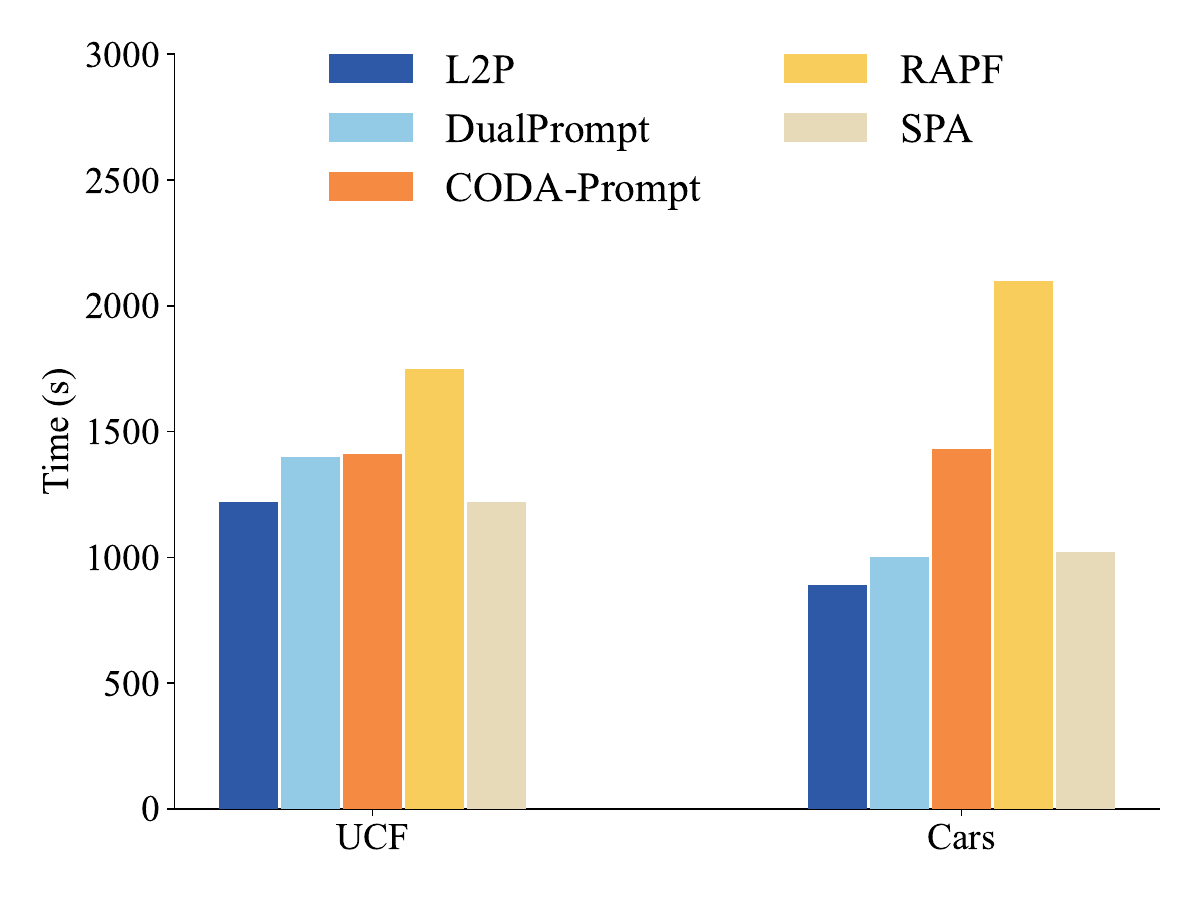}
        \caption{Running time comparison}
        \label{fig:supp-running-time}
    \end{subfigure}
    \hfill
    \begin{subfigure}[t]{0.32\textwidth}
        \centering
        \includegraphics[width=1\linewidth]{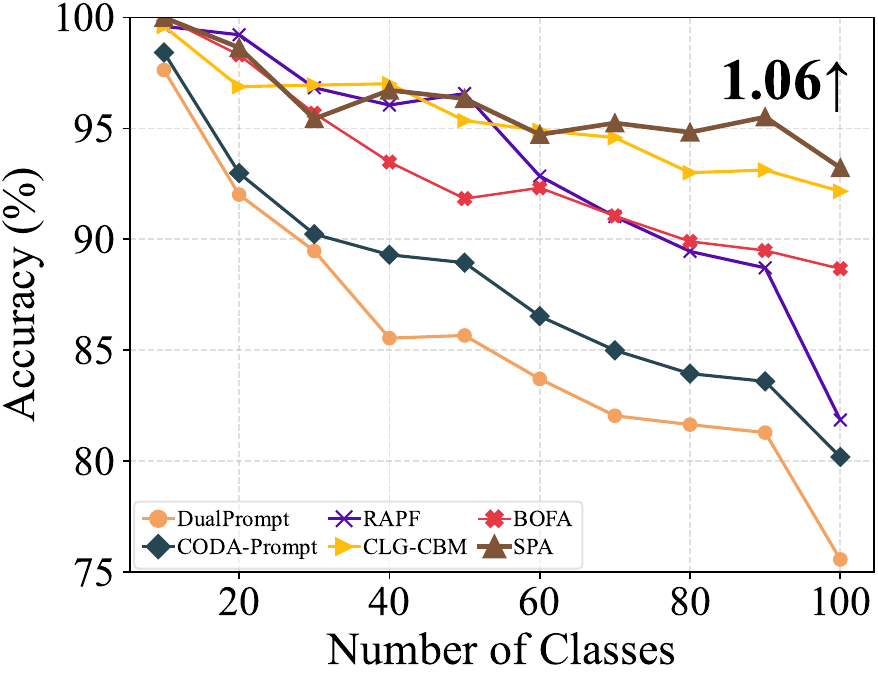}
        \caption{OpenAI pre-trained weight}
        \label{fig:supp-backbone}
    \end{subfigure}
    \hfill
    \begin{subfigure}[t]{0.32\textwidth}
        \centering
        \includegraphics[width=1\linewidth]{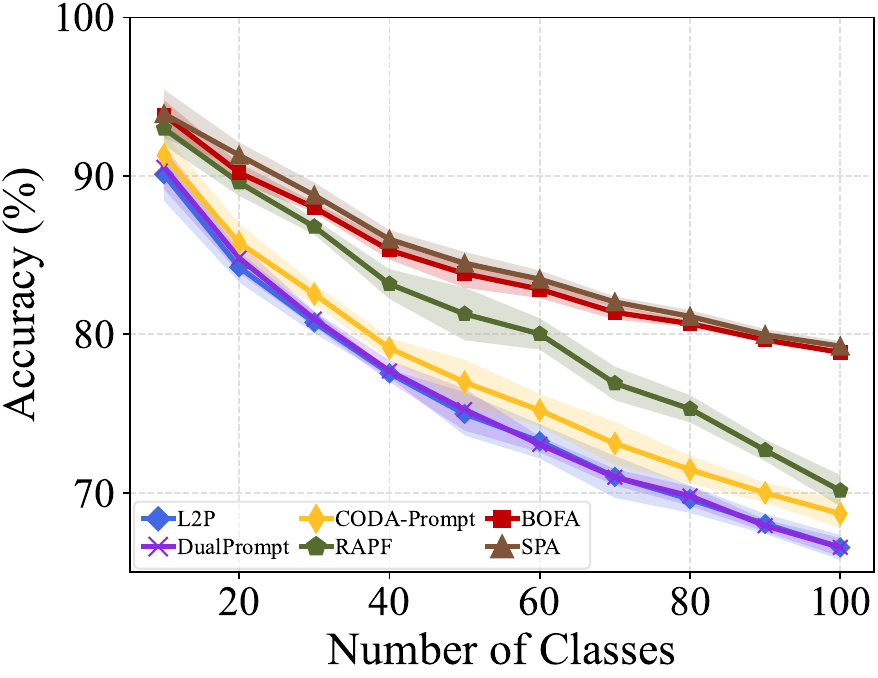}
        \caption{Multiple runs}
        \label{fig:seed}
    \end{subfigure}

    \caption{Running time comparison, OpenAI pre-trained weight results, and multiple runs.}
       \vspace{-6mm}
    \label{fig:-experiments}
\end{figure*}
\subsection{Running Time Comparison}
We further compare the running time of different methods under the same experimental setting. All experiments are conducted on a single NVIDIA 4090 GPU. As shown in Fig.~\ref{fig:supp-running-time}, \name requires less running time than RAPF and CODA-Prompt, and is comparable to prompt-based methods such as L2P and DualPrompt. These results show that \name achieves strong performance without introducing excessive computational overhead.
\subsection{Different Backbones}
In the main paper, we report results using LAION-400M pre-trained CLIP~\cite{ilharco2021openclip}. To further evaluate the generality of \mame, we also conduct experiments with OpenAI pre-trained CLIP~\cite{radford2021learning} on UCF B0 Inc10. As shown in Fig.~\ref{fig:supp-backbone}, \name consistently outperforms compared methods under different pre-trained weights. This indicates that our method is not tied to a specific CLIP initialization and can bring consistent performance gains under different CLIP pre-training settings.
\subsection{Multiple Runs} 
To evaluate the robustness of \mame, we conduct experiments with multiple random class orders on ImageNet-R B0 Inc20. Specifically, we use five random seeds, \textit{i.e.}, 1993, 1994, 1995, 1996, and 1997, to generate different class orders and report the average performance with standard deviation. As shown in Fig.~\ref{fig:seed}, \name consistently achieves strong performance across different runs, showing that it is robust to variations in class order.

\begin{figure*}[t]
 \centering

    \begin{subfigure}[t]{0.48\textwidth}
        \centering
        \includegraphics[width=1\linewidth]{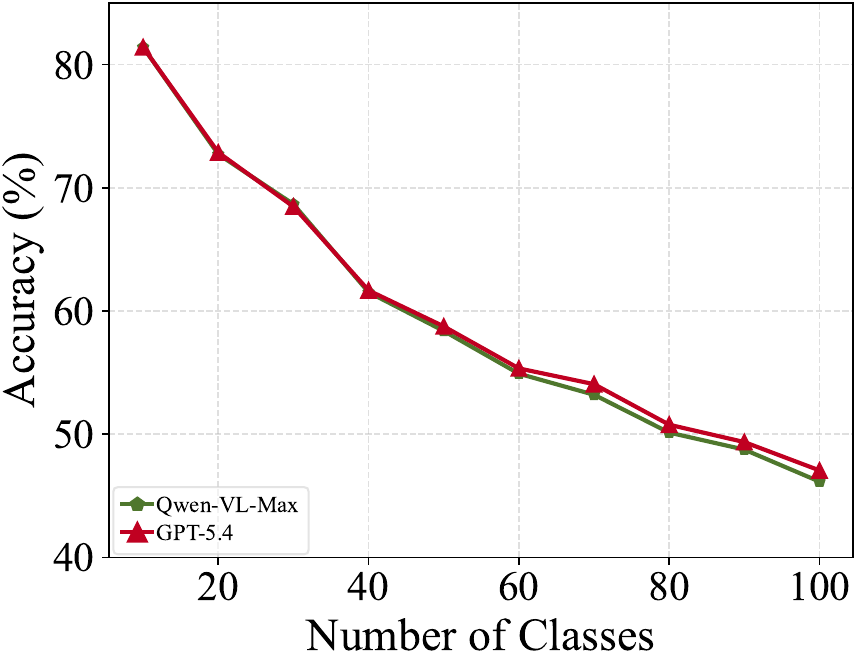}
        \caption{Different LLMs}
        \label{fig:llms}
    \end{subfigure}
    \hfill
    \begin{subfigure}[t]{0.48\textwidth}
        \centering
        \includegraphics[width=1\linewidth]{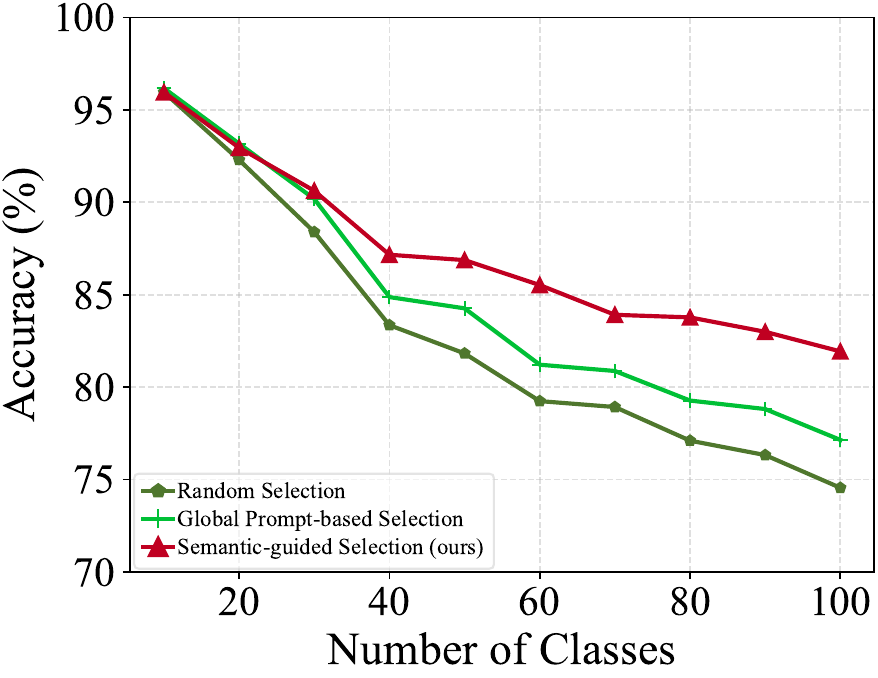}
        \caption{Top-$K$ patch selection strategies}
        \label{fig:topk}
    \end{subfigure}

    \caption{Different LLMs, and Top-$K$ patch selection strategies.}
       \vspace{-2mm}
    \label{fig:experiments}
\end{figure*}

\subsection{Different LLMs}
We further investigate the influence of different LLMs for generating class-wise attribute semantics. Specifically, for fair comparison, we compare GPT-5.4~\cite{achiam2023gpt} with Qwen-VL-Max~\cite{bai2023qwen} under the same visual samples and prompt template on ObjectNet B0 Inc20. As shown in Fig.~\ref{fig:llms}, \name achieves stable performance with both LLMs, indicating that the proposed semantic-guided patch selection is not limited to a specific LLM.

\subsection{Top-K Patch Selection Strategies}
We further compare different top-$K$ patch selection strategies on CUB B0 Inc20 to verify the effectiveness of our semantic-guided patch selection. \textbf{Random Selection} randomly selects $K$ patches from all image patches without using any semantic guidance, while \textbf{Global Prompt-based Selection} ranks all patches according to their similarity to the class-name prompt (\textit{e.g.}, ``a photo of a [CLASS]'') and selects the top-$K$ patches. In contrast, our method (\textit{i.e.}, \textbf{Semantic-guided Selection}) selects patches based on their relevance to class-wise attribute semantics. As shown in Fig.~\ref{fig:topk}, our strategy achieves the best performance, showing that attribute semantics can better identify discriminative local regions and reduce irrelevant background noise.
\subsection{Parameter Robustness}
For the parameter analysis reported in the main paper, we further verify the trends on a validation split. Specifically, we split the original training data into a training subset and a validation subset with a ratio of $4:1$. We conduct hyperparameter sensitivity analysis on the validation subset, and observe trends consistent with those reported in the main paper. These results further support the robustness of \name to different hyperparameter settings.
\subsection{Forgetting Measure}
We evaluate the forgetting degree of different methods using $F_{B}$, which measures how much the performance on previous tasks decreases after learning subsequent tasks. Formally, it is defined as:
\begin{equation}
      F_{B} = \frac{1}{B-1}\sum_{b=1}^{B-1}\max_{l \in \{b, \dots, B-1\}} (\mathcal{A}_{l,b}- \mathcal{A}_{B,b}),
\end{equation}
where $\mathcal{A}_{l,b}$
  denotes the accuracy on task $b$ after learning stage $l$.  A smaller $F_{B}$ indicates better preservation of previously learned knowledge. 
  
\begin{figure*}[t]
   \vspace{-3mm}
 \centering
    \begin{subfigure}[t]{0.48\textwidth}
        \centering
        \includegraphics[width=1\linewidth]{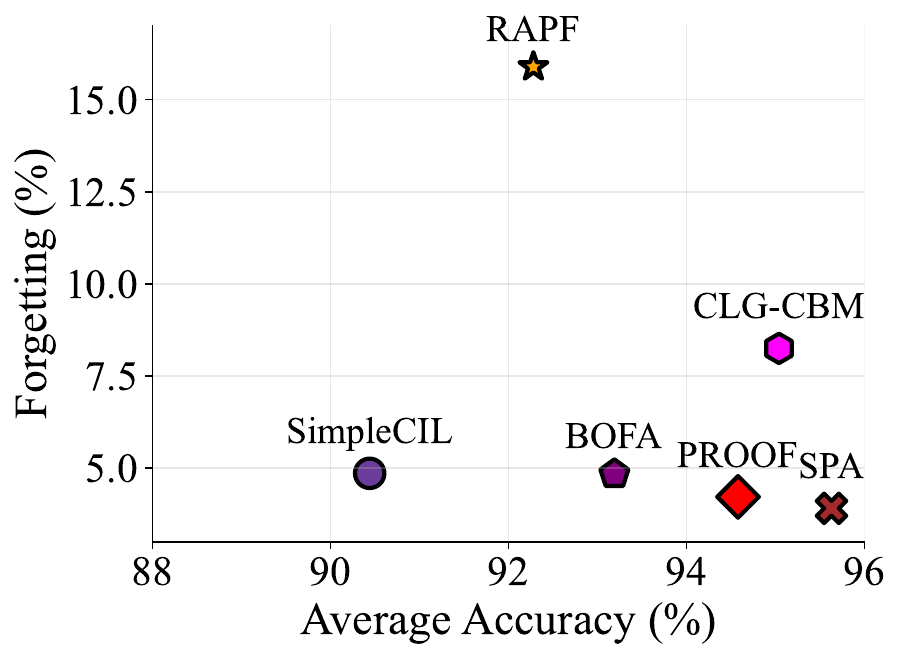}
        \vspace{-1mm}
        \caption{UCF B0 Inc10}
        \label{fig:forgetting_ucf}
    \end{subfigure}
    \hfill
    \begin{subfigure}[t]{0.48\textwidth}
        \centering
        \includegraphics[width=1\linewidth]{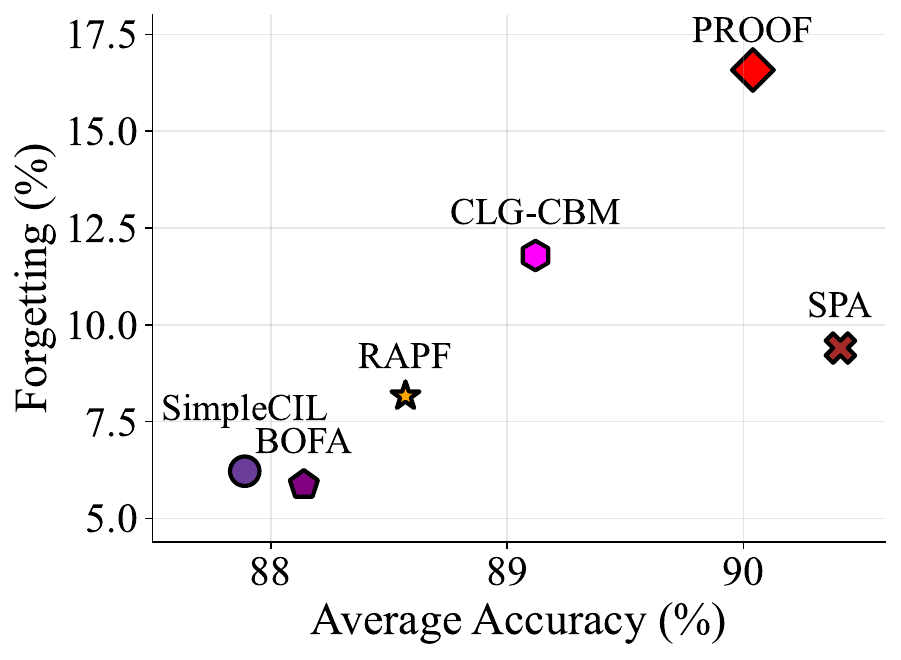}
         \vspace{-1mm}
        \caption{Food B0 Inc10}
        \label{fig:forgetting_food}
    \end{subfigure}
\vspace{-1mm}
    \caption{Accuracy-forgetting trade-off on UCF B0 Inc10 and Food B0 Inc10. The lower-right region indicates higher average accuracy and lower forgetting.}
       \vspace{-4mm}
    \label{fig:forgetting}
\end{figure*}

As shown in Table~\ref{tab:forgetting}, \name maintains relatively low forgetting across different incremental settings, achieving the lowest $F_{B}$
  on UCF B0 Inc10 and both SUN settings, while remaining competitive on Cars. Fig.~\ref{fig:forgetting} further illustrates the trade-off between average accuracy and forgetting, where the lower-right region indicates higher accuracy with less forgetting.  On UCF B0 Inc10, \name is located near this desirable region, achieving the highest average accuracy with the lowest forgetting among compared methods. On Food B0 Inc10, methods such as SimpleCIL and BOFA exhibit relatively small forgetting, but they achieve lower average accuracy than \mame, suggesting a stronger emphasis on stability than plasticity. In contrast, \name achieves the highest average accuracy while maintaining a competitive forgetting measure, demonstrating a favorable stability-plasticity trade-off.
\begin{table*}[t]
	\caption{Forgetting measure $F_{B}$ of different methods under different incremental settings. The best results are highlighted in bold. Lower $F_{B}$	indicates less forgetting.}
	\label{tab:forgetting}
	\vspace{-2mm}
	\centering
	\resizebox{\textwidth}{!}{%
\begin{NiceTabular}{@{} l *{12}{c}}
			\toprule
			\multirow{2}{*}{Method} 
			& \multicolumn{2}{c}{Food} 
			& \multicolumn{2}{c}{Cars} 
            & \multicolumn{2}{c}{UCF} 
			& \multicolumn{2}{c}{SUN} 
			\\
			& B0 Inc10 & B50 Inc10 
			& B0 Inc10 & B50 Inc10 
			& B0 Inc10 & B50 Inc10
          & B0 Inc30 & B150 Inc30 \\
			\midrule
            ZS-CLIP~\citep{radford2021learning} & \textbf{5.22} &\textbf{2.84} &6.41 &3.14 & 10.77& 4.74 &10.16 &4.97\\
            	\rowcolor{gray!10} 
            SimpleCIL~\cite{zhou2025revisiting} & 6.22 &3.76 &5.11 &2.51 & 4.86& \textbf{2.31}&8.12 &3.51\\
                 L2P~\citep{wang2022learning} &24.69 &20.26 &7.45 &4.22 & 8.15& 6.22&22.87 &17.57\\
		\rowcolor{gray!10}
         DualPrompt~\cite{wang2022dualprompt} & 24.28 &20.77 &7.35 &4.71 & 9.26& 7.95&23.54 &17.63\\
            CODA-Prompt~\cite{smith2023coda} & 25.82& 22.41&6.14 &4.59 &9.80 & 9.21&22.71&17.69 \\
            \rowcolor{gray!10}
    RAPF~\cite{huang2024class}  &8.16 & 7.83& 28.27& 29.30& 15.89& 20.50&14.58 &9.36\\
			 CLG-CBM~\citep{yu2025language} & 11.79&9.38 &6.86 & 4.86& 8.25& 6.87 &14.62&12.26\\
             \rowcolor{gray!10}
		PROOF~\citep{zhou2025learning} &16.58 & 16.94& 4.48& 2.30& 4.22& 4.85&18.33 &15.71\\
		BOFA~\citep{li2026bofa} & 5.87&4.08 &\textbf{3.94} & \textbf{2.17}&4.84 &4.13 &8.47&4.16\\
        \hline
	\rowcolor{LightCyan}\name (Ours) 
        		&9.40 &7.93&4.66 &2.22 &\textbf{3.91} & 4.33 &\textbf{6.71}&\textbf{3.31}\\
			\bottomrule
		\end{NiceTabular}
	}	
\end{table*}

\begin{figure*}
	\vspace{-7mm}
	\centering
	  
    \begin{minipage}{\textwidth}
        \centering
        \includegraphics[width=1\textwidth]{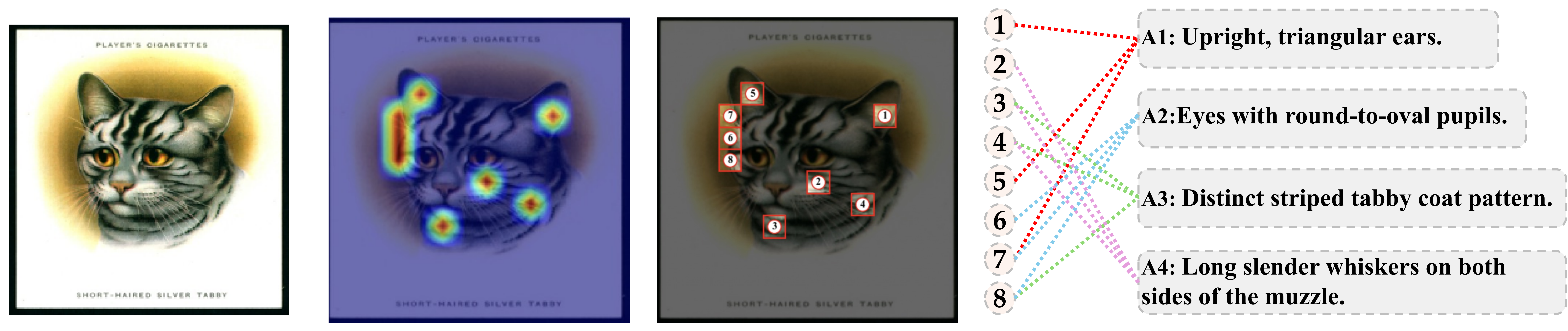} \\
        \label{fig:visual2}
    \end{minipage}

     \begin{minipage}{\textwidth}
        \centering
        \includegraphics[width=1\textwidth]{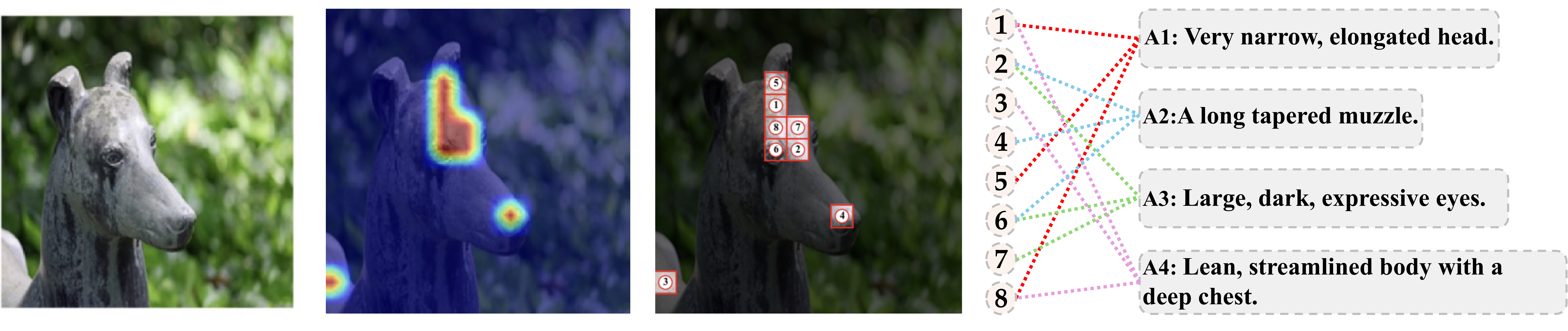} \\
        \label{fig:visual3}
    \end{minipage}

     \begin{minipage}{\textwidth}
        \centering
        \includegraphics[width=1\textwidth]{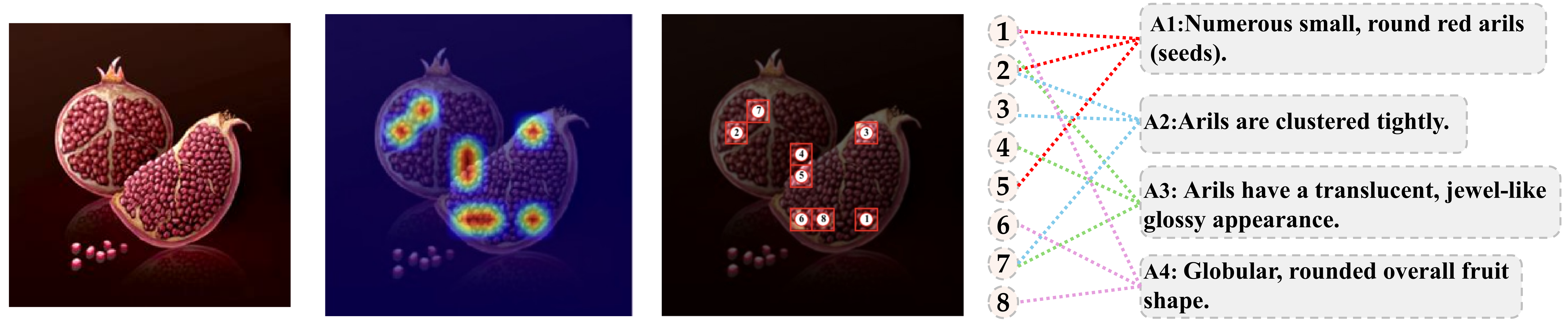} \\
        \label{fig:visual4}
    \end{minipage}

     \begin{minipage}{\textwidth}
        \centering
        \includegraphics[width=1\textwidth]{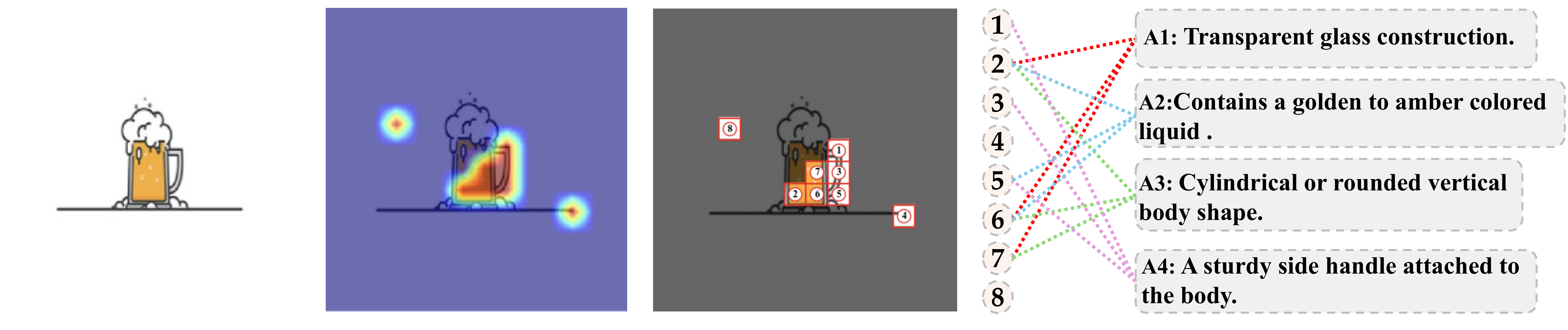} \\
        \label{fig:visual5}
    \end{minipage}
   
	\caption{{\bf Column 1:} Input.  {\bf Column 2:} Response heatmap of discriminative patches.  {\bf Column 3:} Visualization of the top-$8$ discriminative patches selected. {\bf Column 4:} OT-based correspondence graph, where numbered nodes denote the selected patches and attribute nodes denote class-wise attributes. For clarity, only the top-3 correspondences with the highest transport weights are shown.}  
	\vspace{-5mm}
	\label{fig:further-analysis}
\end{figure*}
\section{Visualizations}\label{sec:visualization}
\subsection{More Visualizations} 
Fig.~\ref{fig:further-analysis} provides additional visualizations of the proposed semantic-guided patch selection and OT-based patch-attribute alignment. Across different categories, the response heatmaps show that \name consistently focuses on category-relevant local regions, such as the ears and striped coat of the \textit{cat}, the head and muzzle of the \textit{whippet}, and the clustered arils of the \textit{pomegranate}. In the OT-based correspondence graphs, semantic attributes are structurally associated with their corresponding visual regions. For example, the attribute ``upright, triangular ears'' is linked to patches on the \textit{cat ears}, while ``numerous small, round red arils'' is associated with patches covering the \textit{pomegranate seeds}. These results further demonstrate that \name can establish meaningful token-level relations between local visual patches and textual attributes.
\subsection{Generated class-wise attribute semantics}
In this section, we present examples of class-wise attribute semantics  generated the by GPT-5.4~\cite{achiam2023gpt} on ImageNet-R:\\
\textbf{duck:}
\begin{displayquote}
\vspace{-2mm}
\itshape
	 {\bf Q}: What are the key visual features for identifying a duck in these images? Focus on the most discriminative attributes.\\
	 {\bf A}: {\bf 1.} Relatively short neck connecting the head and body. {\bf 2.} Short, upturned tail at the rear of the body. {\bf 3.} Paired wings along the sides. {\bf 4.} Bills are often colored orange or red.  {\bf 5.} Webbed feet are visible beneath the body. {\bf 6.}$\cdots$
\end{displayquote}
\textbf{snail:}
\begin{displayquote}
\vspace{-2mm}
\itshape
	 	 {\bf Q}: What are the key visual features for identifying a snail in these images? Focus on the most discriminative attributes.\\
	 {\bf A}: {\bf 1.} Distinct rounded coiled shell with a clear spiral pattern. {\bf 2.} Soft, elongated flattened foot extends beneath the shell. {\bf 3.} The shell is proportionally large. {\bf 4.} A head bearing two long upper tentacles.  {\bf 5.} Shells show earthy or golden hues, while the soft body is darker. {\bf 6.}$\cdots$
\end{displayquote}
\textbf{shih\_tzu:}
\begin{displayquote}
\vspace{-2mm}
\itshape
	 	 {\bf Q}: What are the key visual features for identifying a shih\_tzu in these images? Focus on the most discriminative attributes.\\
	 {\bf A}: {\bf 1.} Dense, long, silky coat that covers the body and head. {\bf 2.} Short, pushed-in muzzle with a broad black nose. {\bf 3.} Broad, rounded skull producing a domed forehead. {\bf 4.} Long, heavily coated ears that hang down alongside the face.  {\bf 5.} Small, compact body with short legs. {\bf 6.}$\cdots$
\end{displayquote}
\textbf{pig:}
\begin{displayquote}
\vspace{-2mm}
\itshape
	 	 {\bf Q}: What are the key visual features for identifying a pig in these images? Focus on the most discriminative attributes.\\
	 {\bf A}: {\bf 1.} A short, curly or corkscrew-shaped tail. {\bf 2.} Short, rounded head with a broad snout area. {\bf 3.} Distinct round, flattened snout with two visible nostrils. {\bf 4.} Small eyes set on the sides of the head.  {\bf 5.} Stout, barrel-shaped body that is broad and compact. {\bf 6.}$\cdots$ 
\end{displayquote}
\textbf{mobile\_phone:}
\begin{displayquote}
\vspace{-2mm}
\itshape
	 	 {\bf Q}: What are the key visual features for identifying a mobile\_phone in these images? Focus on the most discriminative attributes.\\
	 {\bf A}: {\bf 1.} Thin, flat rectangular handheld shape. {\bf 2.} Prominent rectangular display that occupies much of the front face. {\bf 3.} Corners and edges are generally rounded. {\bf 4.} Small enough to be held and operated with a single hand.  {\bf 5.} Largely featureless aside from the screen and buttons. {\bf 6.}$\cdots$
\end{displayquote}
\section{Overview of Training Algorithm} \label{sec:alg}
\begin{algorithm}[t]
\caption{Training \name for CIL}
\label{alg:spa}
 \raggedright
 \textbf{Input:}  Current incremental task dataset $\mathcal{D}^b$; Current model: $f(\cdot)$;\\
 \textbf{Output:} Updated model;
\begin{algorithmic}[1]
        \STATE Freeze all previous global and local projectors; \label{line:1}
        \STATE Initialize current global and local projectors;
        \label{line:2}
        \FOR{ each new class   $c$ in $\mathcal{D}^b$} \label{line:3}
            \STATE Extract global visual features via $\bar{g}_i(\cdot)$;\label{line:4}
            \STATE Compute the class prototype $\mathbf{p}_c$ via Eq.~\ref{eq:prototype}; \label{line:5}
            \STATE Construct representative and diverse visual samples $\mathcal{S}_c$ via Eqs.~\ref{eq:vis_pro} and \ref{eq:vis_other}; \label{line:6}
            \STATE Generate class-wise attribute semantics $\mathcal{A}_c$ with GPT-5.4  based on $\mathcal{S}_c$;\label{line:7}
            \STATE Encode $\mathrm{A}_c$ into textual embeddings $\mathcal{T}_c$ via $\bar{g}_t(\cdot)$; \label{line:8}
        \ENDFOR
        
        \FOR{$(\mathbf{x}, y) \in \mathcal{D}^b$}
            \STATE Extract $\mathbf{v}_{\mathrm{cls}}$ and patch-level features $\mathcal{V}$ via $\bar{g}_i(\mathbf{x})$;\label{line:11}
             \STATE  Extract global text embeddings $\mathbf{t}_{\mathrm{eos}}^y$ via $\bar{g}_t(\cdot)$;\label{line:12}
            \STATE Compute adapted global and local features with task-specific projectors via Eq.~\ref{eq:12}; \label{line:14}
            \STATE Select top-$K$ discriminative patches guided by $\mathcal{T}_y$ via Eq.~\ref{eq:dis_set};\label{line:13}
            \STATE Perform patch-level alignment via optimal transport in Eq.~\ref{eq:transport};\label{line:16}
            \STATE Compute the global loss $\mathcal{L}_g$ via Eq.~\ref{eq:unit-loss};
           \label{line:15}
            \STATE Compute the local loss $\mathcal{L}_l$ via Eq.~\ref{eq:local-loss};\label{line:17}
            \STATE Compute the total loss $\mathcal{L}$ via Eq.~\ref{eq:total-loss};
            update the model; \label{line:18}
        \ENDFOR
        \RETURN the updated model.
\end{algorithmic}
\end{algorithm}
The training process of SPA is detailed in Algorithm~\ref{alg:spa}. At the beginning of the $b$-th incremental task, we freeze the previously learned visual and textual projectors for both global and local branches, and initialize the current ones (Line~\ref{line:1} to Line~\ref{line:2}). For each new class $c \in Y_b$, \name first computes the prototype $\mathbf{p}_c$ (Line~\ref{line:5}), constructs the representative and diverse visual sample set $\mathcal{S}_c$ (Line~\ref{line:6}), and generates class-wise attribute semantics  $\mathrm{A}_c$ with GPT-5.4 based on $\mathcal{S}_c$ (Line~\ref{line:7}). During training, we extract the global visual feature $\mathbf{v}_{\mathrm{cls}}$, patch-level feature set $\mathcal{V}$, and global text embeddings $\mathbf{t}_{\mathrm{eos}}^y$ using frozen CLIP encoders (Line~\ref{line:11} to Line~\ref{line:12}). Then, the current projectors and frozen historical projectors are cumulatively applied to obtain adapted global and local features. Based on the adapted features (Line~\ref{line:14}), \name selects top-$K$ discriminative patches guided by $\mathcal{T}_y$ (Line~\ref{line:13}), which are further aligned with textual attributes via optimal transport (Line~\ref{line:16}). Finally, \name jointly optimizes the global loss $\mathcal{L}_g$ and local loss $\mathcal{L}_l$ to update only the current task projectors (Line~\ref{line:15} to Line~\ref{line:18}). The final updated model is then
returned as the output of the training process.
\begin{table*}[t]
	\caption{Comparison of the average and last performance of different methods. The best results are highlighted in bold. \textbf{All methods are initialized from the same pre-trained CLIP backbone.}}\label{tab:full_results}
	\vspace{-2mm}
	\centering
	\resizebox{\textwidth}{!}{%
	\begin{NiceTabular}{@{} l *{15}{c}}
			\toprule
			\multicolumn{1}{c}{\multirow{3}{*}{Method}}
			&
			\multicolumn{4}{c}{Aircraft }   & 
			\multicolumn{4}{c}{CIFAR100 }	&	
			\multicolumn{4}{c}{Cars }   
			\\ 
			& 
			\multicolumn{2}{c}{B0 Inc10}   & 
			\multicolumn{2}{c}{B50 Inc10}	&		
			\multicolumn{2}{c}{B0 Inc10}   & 
			\multicolumn{2}{c}{B50 Inc10}	& 
			\multicolumn{2}{c}{B0 Inc10}   & 
			\multicolumn{2}{c}{B50 Inc10}	& 
			\\  
			& 
			{$\bar{\mathcal{A}}$} & ${\mathcal{A}_B}$  
			& {$\bar{\mathcal{A}}$} & ${\mathcal{A}_B}$
			& {$\bar{\mathcal{A}}$} & ${\mathcal{A}_B}$ 
			&  {$\bar{\mathcal{A}}$} & ${\mathcal{A}_B}$  
			& {$\bar{\mathcal{A}}$} & ${\mathcal{A}_B}$
			& {$\bar{\mathcal{A}}$} & ${\mathcal{A}_B}$ 
			\\
			\midrule
   ZS-CLIP~\cite{radford2021learning} &26.66 & 17.22 & 21.70 & 17.22& 81.81 & 71.38& 76.49 & 71.38& 82.60 & 76.37& 78.32 & 76.37\\
			\rowcolor{gray!10} SimpleCIL~\cite{zhou2025revisiting} &59.24 & 48.09 & 53.05 & 48.09 & 84.15 & 76.63& 80.20 & 76.63& 92.04 & 86.85 & 88.96 & 86.85\\
			L2P~\cite{wang2022learning}  &47.19 & 28.29 &44.07&32.13& 82.74 & 73.03& 81.14 & 73.61& 76.63 & 61.82& 76.37 & 65.64 \\
   \rowcolor{gray!10}
			DualPrompt~\cite{wang2022dualprompt}  & 44.30& 25.83 &46.07&33.57 & 81.63 & 72.44& 80.12 & 72.57& 76.26 & 62.94& 76.88 & 67.55 \\
			CODA-Prompt~\cite{smith2023coda}  & 45.98 & 27.69 & 45.14 & 32.28& 82.43 & 73.43& 78.69 & 71.58& 80.21 & 66.47& 75.06 & 64.19 \\
   \rowcolor{gray!10}
			RAPF~\cite{huang2024class}   &  50.38  & 23.61 &  40.47 &  25.44 & 86.14 & 78.04 & 82.17 &  77.93  & 82.89 & 62.85 &  75.87 & 63.19\\
   CLG-CBM~\citep{yu2025language}  & 66.05   &55.93 &59.25 & 55.39 &  86.58 & 80.15  &83.59  &79.28 &93.25  & 88.76    & 90.11  &88.19 \\
   \rowcolor{gray!10}
   PROOF~\citep{zhou2025learning}  &  63.81  & 56.14 &59.47 & 57.10& 86.77  &79.11   &83.32  & 79.73 &90.74  & 86.51    & 88.00  & 85.58\\
   BOFA~\citep{li2026bofa}  &70.96    &60.43  &66.09 &61.36 &  86.07 &  79.19 &  83.02&  79.44& 94.21 &  90.20 & 92.13  &90.50 \\
   \hline
	\rowcolor{LightCyan}\name (Ours) &\bf 71.57 & \bf61.51 & \bf66.82 & \bf 63.01 & \bf88.53 & \bf 81.81 &  \bf85.01 & \bf 81.60 &  \bf  94.43& \bf 90.91 & \bf 92.33 &  \bf 91.43 \\
		\end{NiceTabular}
	}	
	\resizebox{\textwidth}{!}{%
		\begin{NiceTabular}{@{} l *{15}{c}}
			\toprule
			\multicolumn{1}{c}{\multirow{3}{*}{Method}}
			& 
			\multicolumn{4}{c}{ImageNet-R }   & 
			\multicolumn{4}{c}{CUB }	&	\multicolumn{4}{c}{UCF }   
			\\ 
			& 
			\multicolumn{2}{c}{B0 Inc20}   & 
			\multicolumn{2}{c}{B100 Inc20}	&	\multicolumn{2}{c}{B0 Inc20}   & 
			\multicolumn{2}{c}{B100 Inc20}	& 
			\multicolumn{2}{c}{B0 Inc10}   & 
			\multicolumn{2}{c}{B50 Inc10}	& 
			\\  
			& 
			{$\bar{\mathcal{A}}$} & ${\mathcal{A}_B}$  
			& {$\bar{\mathcal{A}}$} & ${\mathcal{A}_B}$
			& {$\bar{\mathcal{A}}$} & ${\mathcal{A}_B}$ 
			&  {$\bar{\mathcal{A}}$} & ${\mathcal{A}_B}$  
			& {$\bar{\mathcal{A}}$} & ${\mathcal{A}_B}$
			& {$\bar{\mathcal{A}}$} & ${\mathcal{A}_B}$ 
			\\
			\midrule
			ZS-CLIP~\cite{radford2021learning} &83.37 & 77.17& 79.57 & 77.17 & 74.38 & 63.06& 67.96 & 63.06& 75.50 & 67.64& 71.44 & 67.64\\
    \rowcolor{gray!10}
			SimpleCIL~\cite{zhou2025revisiting} & 81.06 & 74.48& 76.84 & 74.48& 83.81 & 77.52& 79.75 & 77.52& 90.44 & 85.68& 88.12 & 85.68\\
			L2P~\cite{wang2022learning}  &75.97 & 66.52 & 72.82 & 66.77&   70.87&57.93 & 75.64 &66.12 & 86.34 & 76.43& 83.95 & 76.62 \\
    \rowcolor{gray!10}
			DualPrompt~\cite{wang2022dualprompt}  &76.21 & 66.65 & 73.22 & 67.58&69.89 &57.46 & 74.40 &64.84 & 85.21 & 75.82& 84.31 & 76.35 \\
			CODA-Prompt~\cite{smith2023coda}  & 77.69 & 68.95 & 73.71 & 68.05& 73.12&62.98 &73.95&62.21 & 87.76 & 80.14&83.04 & 75.03 \\
    \rowcolor{gray!10}
			RAPF~\cite{huang2024class}  & 81.26  & 70.48 & 76.10 & 70.23 &  79.09 & 62.77& 72.82 & 62.93 & 92.28 & 80.33&90.31 & 81.55\\
			
             CLG-CBM~\citep{yu2025language}  &  84.64  & 78.50& 81.46& 77.88 &  85.37 & 78.24  &77.74  &76.97 & 95.04&  91.36   & 94.17  &91.85 \\
     \rowcolor{gray!10}
   PROOF~\citep{zhou2025learning}  &83.84    &78.40  &81.20 & 78.92&82.31   & 76.64  &79.20  &76.37  & 94.58 & 91.10    &  93.58 &90.91 \\
   BOFA~\citep{li2026bofa}   & 84.53   &78.77  & 81.60 &79.12 & 86.66  &80.58   & 83.18 & 80.79 & 93.19 &  88.71  &92.60   &89.43 \\
 \hline
  \rowcolor{LightCyan}\name (Ours)  & \bf 85.63 & \bf 79.08 & \bf 82.50 & \bf 79.50 & \bf87.17 & \bf81.93 &  \bf 84.43& \bf 82.23&  \bf 95.63 & \bf 92.38  & \bf 95.43 &  \bf 93.48  \\
		\end{NiceTabular}
	}
	
	\resizebox{\textwidth}{!}{%
		\begin{NiceTabular}{@{} l *{15}{c}}
			\toprule
			\multicolumn{1}{c}{\multirow{3}{*}{Method}}
			&
			\multicolumn{4}{c}{SUN }   & 
			\multicolumn{4}{c}{Food }	&	\multicolumn{4}{c}{ObjectNet }   
			\\ 
			& 
			\multicolumn{2}{c}{B0 Inc30}   & 
			\multicolumn{2}{c}{B150 Inc30}	&		\multicolumn{2}{c}{B0 Inc10}   & 
			\multicolumn{2}{c}{B50 Inc10}	& 
			\multicolumn{2}{c}{B0 Inc20}   & 
			\multicolumn{2}{c}{B100 Inc20}	& 
			\\  
			& 
			{$\bar{\mathcal{A}}$} & ${\mathcal{A}_B}$  
			& {$\bar{\mathcal{A}}$} & ${\mathcal{A}_B}$
			& {$\bar{\mathcal{A}}$} & ${\mathcal{A}_B}$ 
			&  {$\bar{\mathcal{A}}$} & ${\mathcal{A}_B}$  
			& {$\bar{\mathcal{A}}$} & ${\mathcal{A}_B}$
			& {$\bar{\mathcal{A}}$} & ${\mathcal{A}_B}$ 
			\\
			\midrule
   ZS-CLIP~\cite{radford2021learning} &79.42 & 72.11& 74.95 & 72.11& 87.86 & 81.92& 84.75 & 81.92& 38.43 & 26.43 & 31.12 & 26.43\\	
    \rowcolor{gray!10}
			SimpleCIL~\cite{zhou2025revisiting} & 82.13 & 75.58& 78.62 & 75.58& 87.89 & 81.65& 84.73 & 81.65& 52.06 & 40.13& 45.11 & 40.13\\
			L2P~\cite{wang2022learning} &82.82 & 74.54 & 79.57 & 73.10 & 85.66 & 77.33& 80.42 & 73.13& 51.40 & 39.39& 48.91 & 42.83 \\
    \rowcolor{gray!10}
			DualPrompt~\cite{wang2022dualprompt} & 82.46 & 74.40 & 79.37 & 73.02& 84.92 &77.29& 80.00 & 72.75& 52.62 & 40.72& 49.08 & 42.92 \\
			CODA-Prompt~\cite{smith2023coda} &   83.34 & 75.71 & 80.38 & 74.17& 86.18 & 78.78& 80.98 & 74.13& 46.49 & 34.13& 40.57 & 34.13 \\
    \rowcolor{gray!10}
			RAPF~\cite{huang2024class}  & 82.13 & 72.47 & 78.04 & 73.10 & 88.57 & 81.15& 85.53 & 81.17&  48.67 & 27.43 & 39.28 &  28.73 \\
			
            CLG-CBM~\citep{yu2025language}  &  84.85  & 78.09&81.58 & 77.79 &  89.12 &83.05   & 86.76 &83.85 &58.53 &45.11   & 49.80  &43.56 \\
  \rowcolor{gray!10}
   PROOF~\citep{zhou2025learning}  & 83.89   &77.25  & 80.15&76.54 &  90.04 & 84.73  &87.52  & 84.74 & 56.07 &  43.69   & 48.90  & 43.62\\
   BOFA~\citep{li2026bofa}  & 84.38   &77.60  & 81.34 &77.93 &88.14   & 82.08  & 85.97 &  82.84& 59.21 & 46.95    & 51.89  &46.76 \\ 
   \hline
    \rowcolor{LightCyan}\name (Ours) & \bf 86.91 & \bf 80.77 & \bf83.57 & \bf80.74 & \bf90.41  & \bf 84.81&  \bf 87.80& \bf 84.95&  \bf 59.98  & \bf 47.06 & \bf 53.28 &  \bf 47.59  \\
			\bottomrule
	\end{NiceTabular}
	}
    \vspace{-3mm}
\end{table*}
\section{Introduction About Compared Methods} \label{sec:a2}
In this section, we briefly introduce the compared methods used in our experiments. All methods are initialized from the same pre-trained CLIP backbone.\\
\textbf{ZS-CLIP~\citep{radford2021learning}:} This baseline directly performs prediction with the frozen CLIP without any training or parameter updates. It serves as a zero-shot performance reference for pre-trained CLIP in the incremental setting.\\
\textbf{SimpleCIL~\cite{zhou2025revisiting}:} This baseline freezes the pre-trained visual encoder and performs classification with the extracted visual representations, without using the text encoder. It provides a simple yet strong reference for CIL.\\
\textbf{L2P~\citep{wang2022learning}:} This method learns the prompt pool and retrieves suitable prompts according to the input image to adapt a frozen visual backbone. Since it is designed for visual prompt learning, it does not exploit the CLIP text encoder or textual semantics.\\
\textbf{DualPrompt~\citep{wang2022dualprompt}:} Built upon L2P, this method introduces two types of prompts, \textit{i.e.}, general prompts and expert prompts, to capture task-shared and task-specific knowledge, respectively. Similar to L2P, it only adapts the visual branch and does not use the text encoder.\\
\textbf{CODA-Prompt~\citep{smith2023coda}:}  This method learns decomposed attention-based prompts and dynamically composes prompts for each instance. It only uses the visual branch of CLIP.\\
\textbf{RAPF~\citep{huang2024class}:} This method is a CLIP-based CIL approach that leverages textual knowledge to adaptively adjust class prototypes and further applies parameter fusion to reduce interference between incremental tasks.\\
 \textbf{CLG-CBM~\citep{yu2025language}:} This method builds a language-guided concept bottleneck model for continual learning. It maps visual representations into interpretable concept spaces and leverages textual concepts to improve both recognition performance and prediction interpretability.\\
\textbf{PROOF~\citep{zhou2025learning}:} This method introduces task-specific projection layers on frozen CLIP encoders and a cross-modal fusion layer to integrate visual and textual knowledge. By aggregating the learned projections across incremental tasks, it adapts CLIP to new classes while preserving previously acquired knowledge.\\
 \textbf{BOFA~\citep{li2026bofa}:} This method proposes bridge-layer orthogonal low-rank fusion for CLIP-based CIL. It introduces lightweight low-rank updates and constrains them to reduce interference between old and new tasks, improving the stability of incremental adaptation.
\section{Full Results} \label{sec:full-results}

\begin{figure*} 
\vspace{-7mm}
	\centering
	\begin{subfigure}{0.32\textwidth}
		\includegraphics[width=1\columnwidth]{figure/aircraft_10.pdf}
		\caption{Aircraft Base0 Inc10}
	\end{subfigure}
	\hfill
	\begin{subfigure}{0.32\linewidth}
		\includegraphics[width=1\linewidth]{figure/cifar_10.pdf}
		\caption{CIFAR100 Base0 Inc10}
	\end{subfigure}
	\hfill
	\begin{subfigure}{0.32\linewidth}
		\includegraphics[width=1\linewidth]{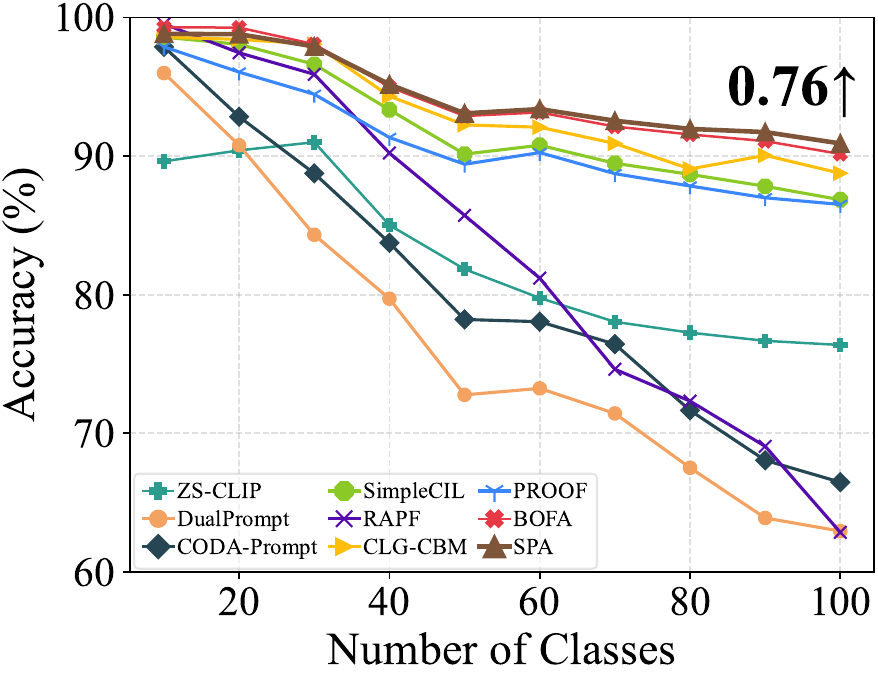}
		\caption{Cars Base0 Inc10}
	\end{subfigure}
	\\
	\begin{subfigure}{0.32\linewidth}
		\includegraphics[width=1\linewidth]{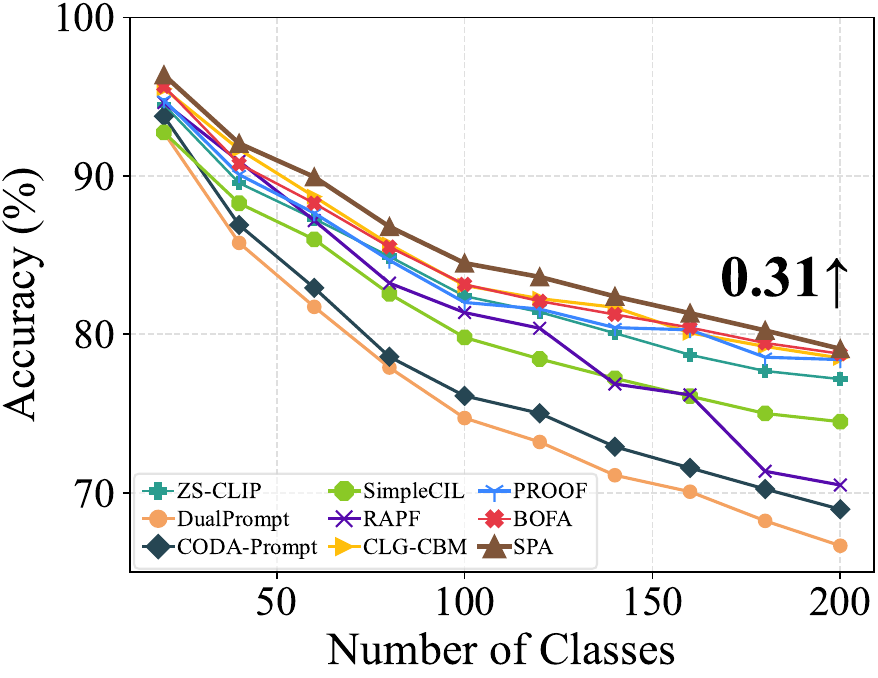}
		\caption{ImageNet-R Base0 Inc20}
	\end{subfigure}
	\hfill
	\begin{subfigure}{0.32\linewidth}
		\includegraphics[width=1\linewidth]{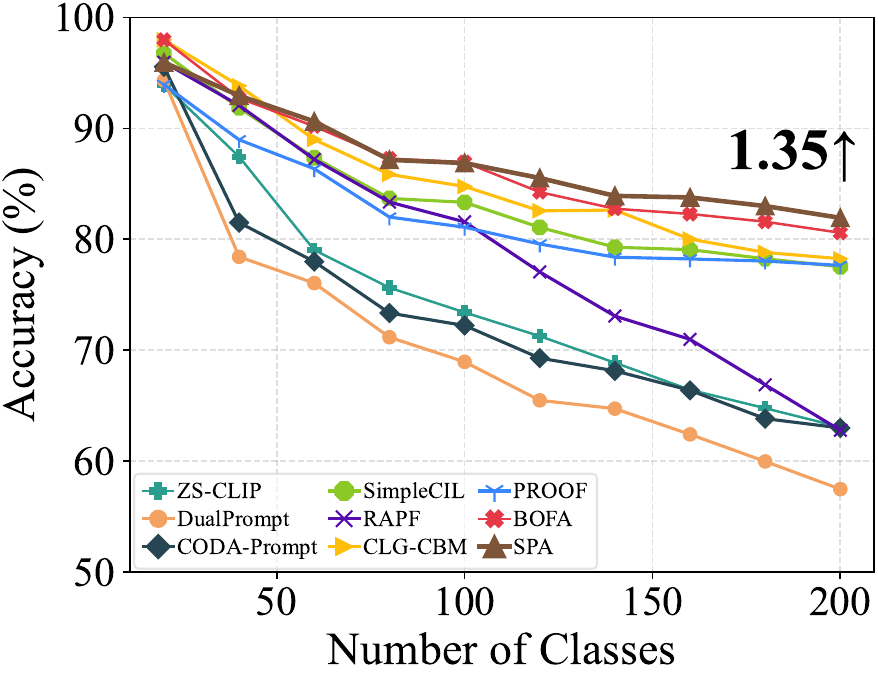}
		\caption{CUB Base0 Inc20}
	\end{subfigure}
	\hfill
	\begin{subfigure}{0.32\linewidth}
		\includegraphics[width=1\columnwidth]{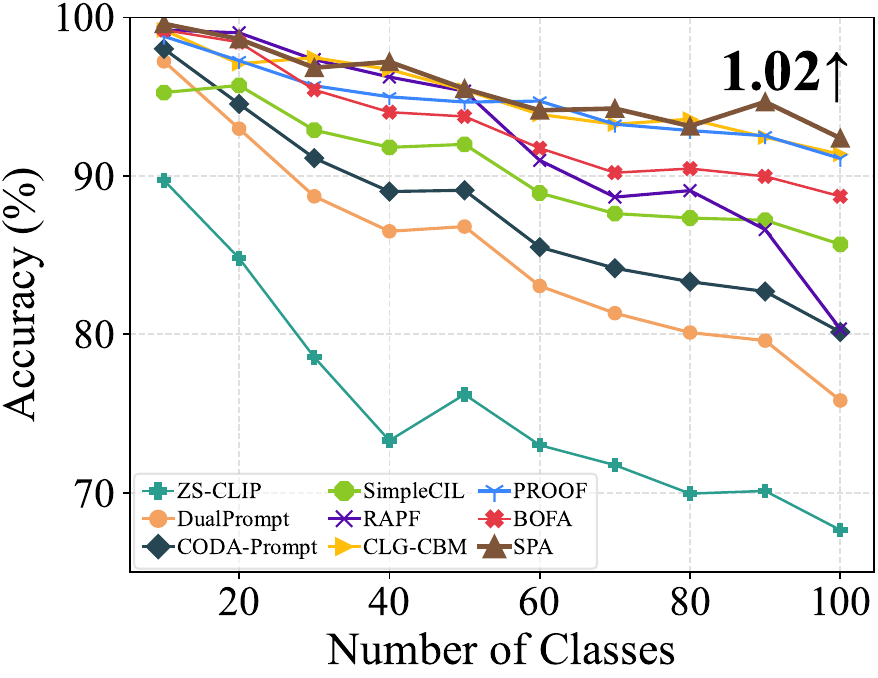}
		\caption{UCF Base0 Inc10}
	\end{subfigure}
	\\
	\begin{subfigure}{0.32\linewidth}
		\includegraphics[width=1\linewidth]{figure/sun_30.pdf}
		\caption{SUN Base0 Inc30}
	\end{subfigure}
	\hfill
	\begin{subfigure}{0.32\linewidth}
		\includegraphics[width=1\linewidth]{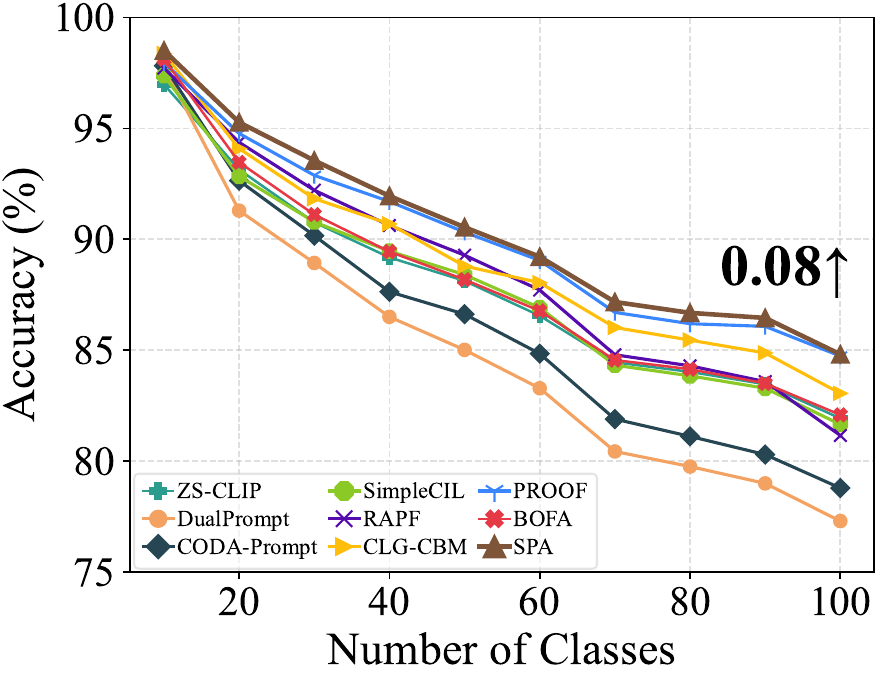}
		\caption{Food Base0 Inc10}
	\end{subfigure}
	\hfill
	\begin{subfigure}{0.32\linewidth}
		\includegraphics[width=1\columnwidth]{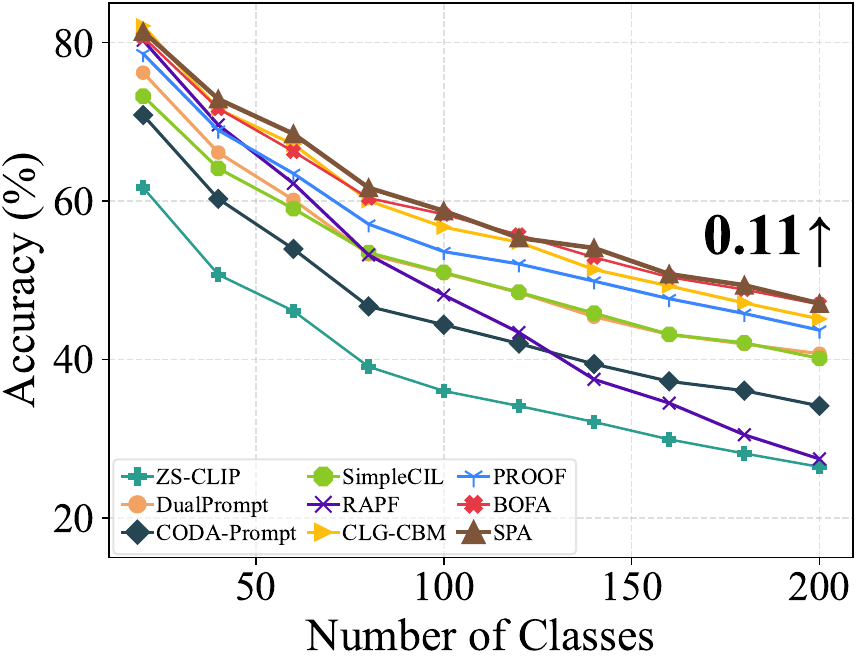}
		\caption{ObjectNet Base0 Inc20}
	\end{subfigure}
	\vspace{-1mm}
	\caption{Incremental performance of different methods on the B0 setting. We report the performance gap after the last incremental stage of \name and the runner-up method at the end of the line.}
		\vspace{-4mm}
	\label{fig:more-supp-benchmark-b0}
\end{figure*}

\begin{figure*}
	\centering
	\begin{subfigure}{0.32\linewidth}
		\includegraphics[width=1\columnwidth]{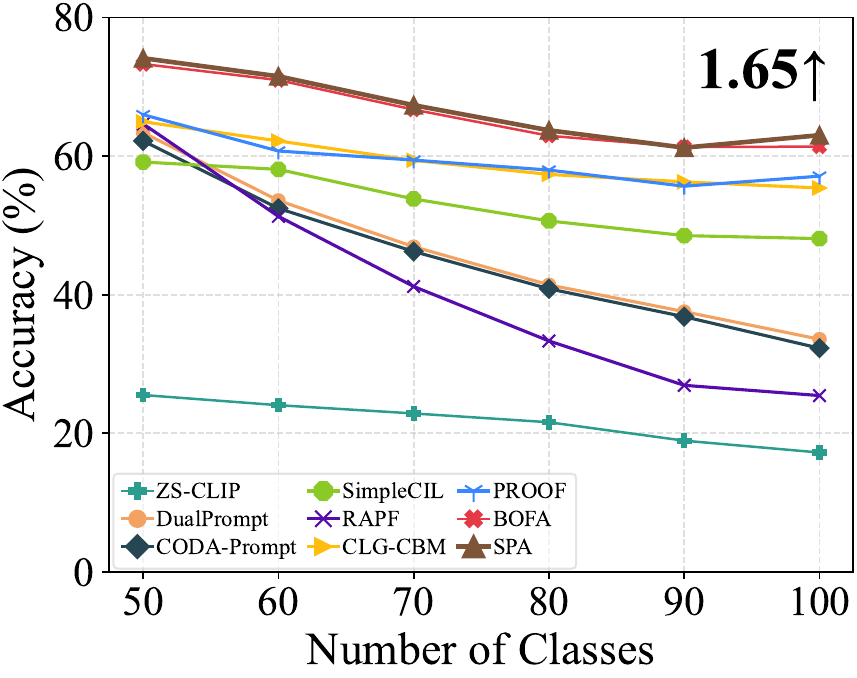}
		\caption{Aircraft Base50 Inc10}
		\label{fig}
	\end{subfigure}
	\hfill
	\begin{subfigure}{0.32\linewidth}
		\includegraphics[width=1\linewidth]{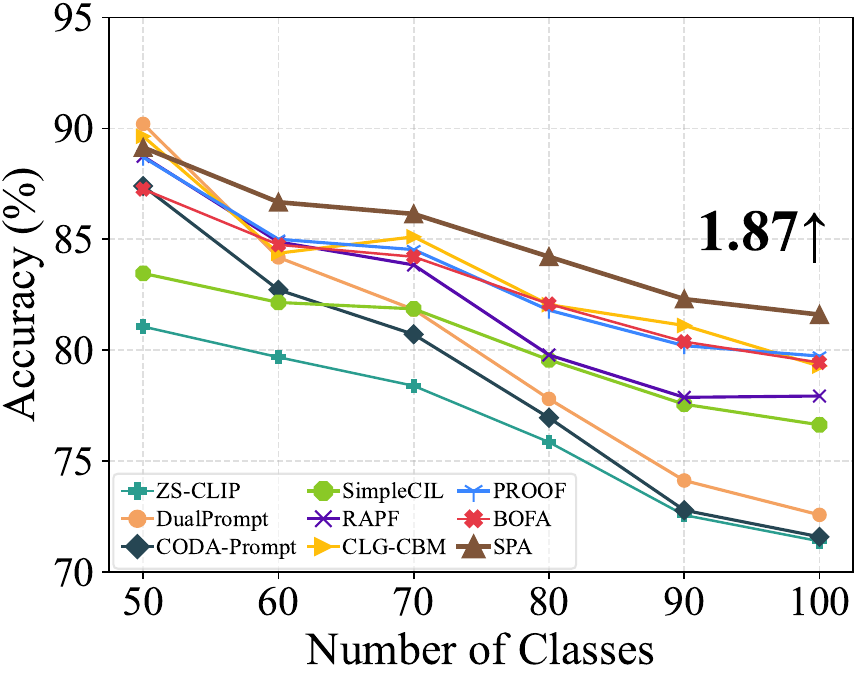}
		\caption{CIFAR100 Base50 Inc10}
		\label{fig:benchmark-cifar50}
	\end{subfigure}
	\hfill
	\begin{subfigure}{0.32\linewidth}
		\includegraphics[width=1\linewidth]{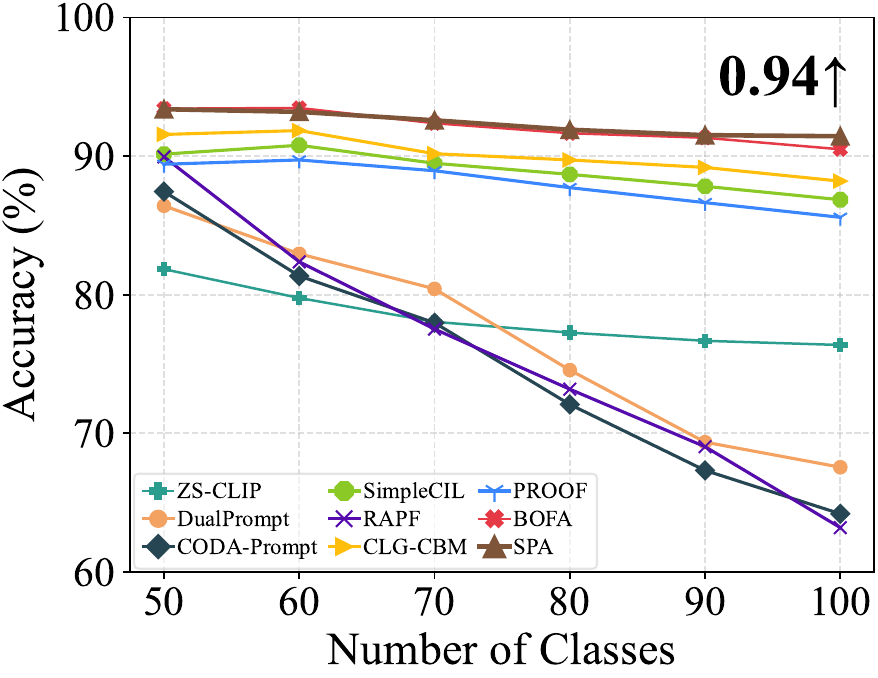}
		\caption{Cars Base50 Inc10}
		\label{fig:benchmark-cars50}
	\end{subfigure}
	\\
	\begin{subfigure}{0.32\linewidth}
		\includegraphics[width=1\linewidth]{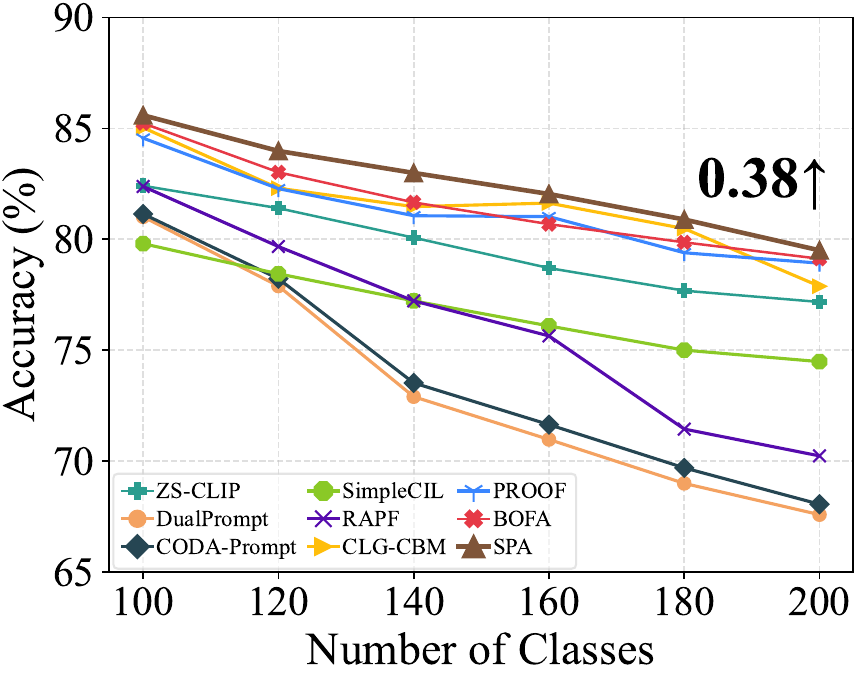}
		\caption{ImageNet-R Base100 Inc20}
		\label{fig:benchmark-imagenetr100}
	\end{subfigure}
	\hfill
	\begin{subfigure}{0.32\linewidth}
		\includegraphics[width=1\linewidth]{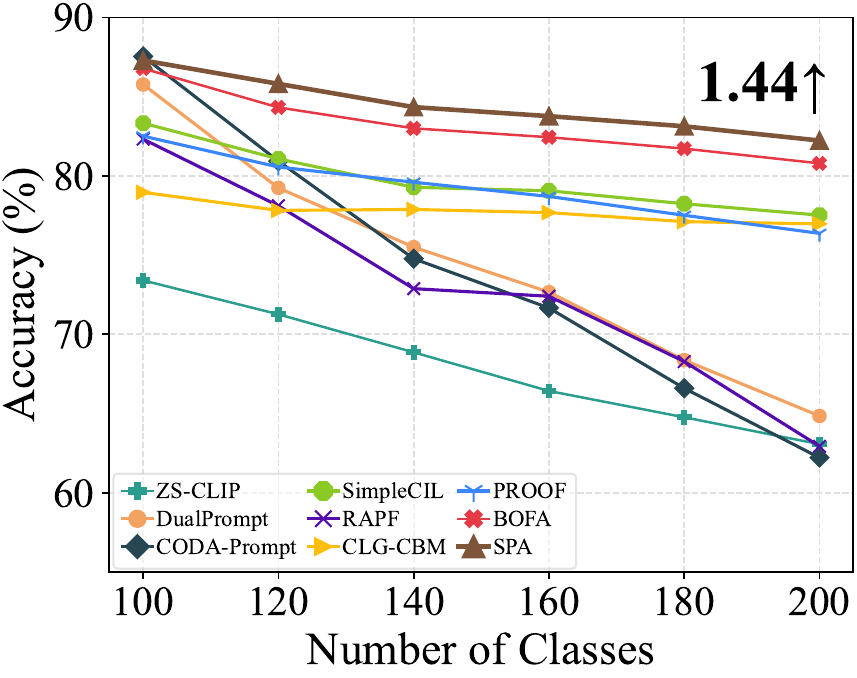}
		\caption{CUB Base100 Inc20}
		\label{fig:benchmark-cub100}
	\end{subfigure}
	\hfill
	\begin{subfigure}{0.32\linewidth}
		\includegraphics[width=1\columnwidth]{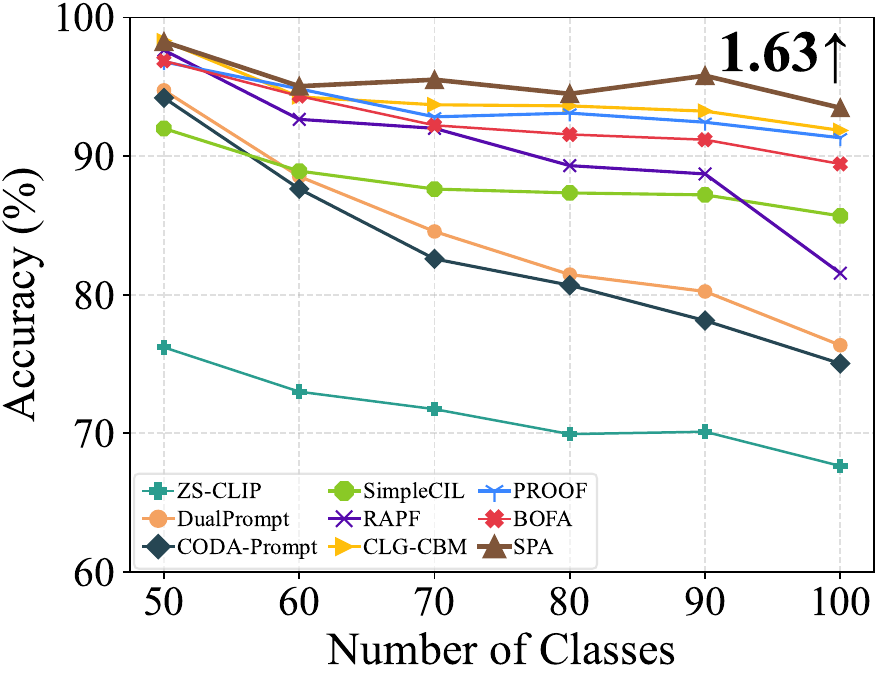}
		\caption{UCF Base50 Inc10}
		\label{fig:benchmark-ucf50}
	\end{subfigure}
	\\
	\begin{subfigure}{0.32\linewidth}
		\includegraphics[width=1\linewidth]{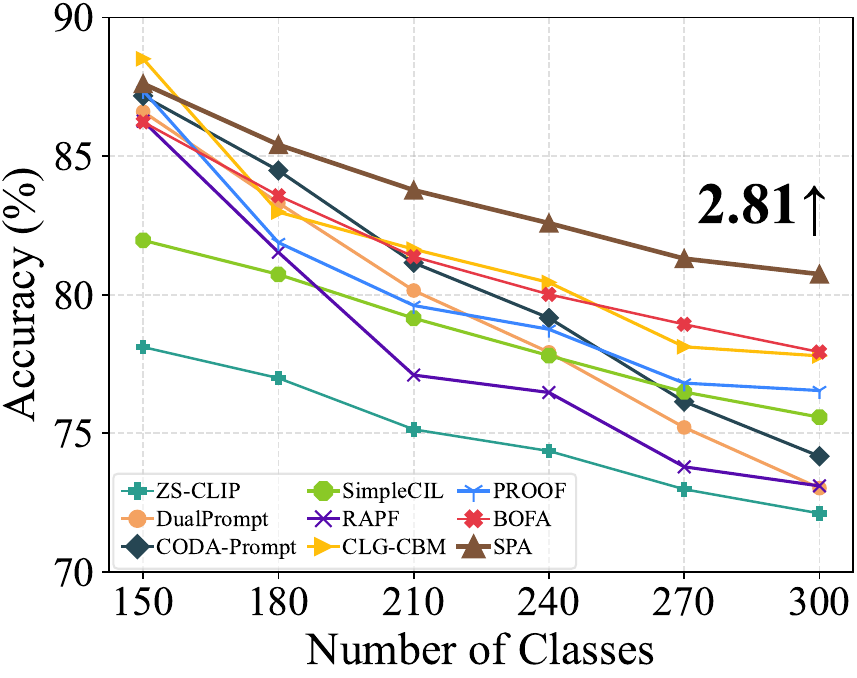}
		\caption{SUN Base150 Inc30}
		\label{fig:benchmark-sun150}
	\end{subfigure}
	\hfill
	\begin{subfigure}{0.32\linewidth}
		\includegraphics[width=1\linewidth]{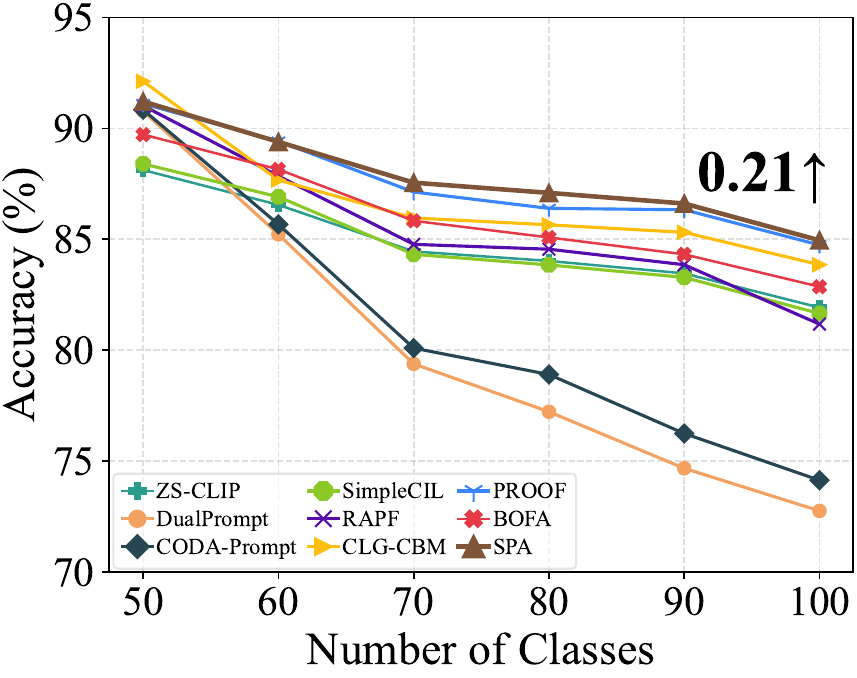}
		\caption{Food Base50 Inc10}
		\label{fig:benchmark-food50}
	\end{subfigure}
	\hfill
	\begin{subfigure}{0.32\linewidth}
		\includegraphics[width=1\columnwidth]{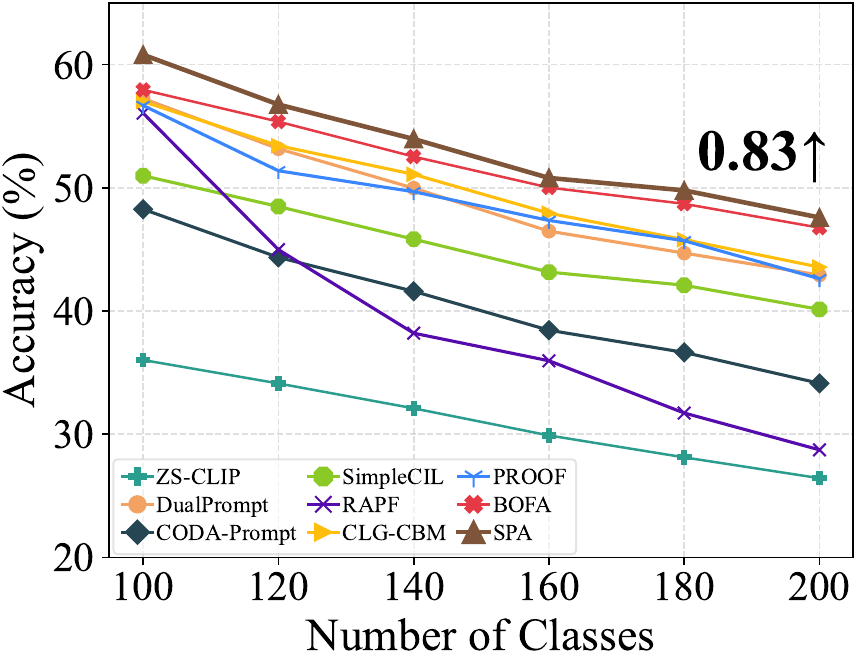}
		\caption{ObjectNet Base100 Inc20}
		\label{fig:benchmark-objectnet100}
	\end{subfigure}
	\caption{Incremental performance of different methods on a half-base setting. We report the performance gap after the last incremental stage of \name and the runner-up method at the end of the line.}
	\label{fig:supp-benchmark-b50}
\end{figure*}

In this section, we provide the complete incremental performance results of different methods in Table~\ref{tab:full_results}. We report the incremental performance curves under the B0 setting in Fig.~\ref{fig:more-supp-benchmark-b0} and the half-base setting in Fig.~\ref{fig:supp-benchmark-b50}. As shown in these results, \name consistently maintains strong performance across different datasets and data splits, demonstrating its effectiveness in CLIP-based CIL.

\section{Broader Impacts} \label{sec:impact}
This work advances CLIP-based class-incremental learning, benefiting applications that require efficient adaptation to evolving visual categories, such as long-term visual recognition and resource-constrained model updating. Since \name adapts CLIP with lightweight modules instead of training from scratch, it may also reduce computational costs and energy consumption. However, the method relies on LLM-generated semantics, which may contain biases or inaccurate descriptions. Practical use should validate generated semantics and avoid privacy-sensitive or ethically sensitive applications.

\end{document}